\documentclass{article}

\usepackage[final, nonatbib]{neurips_2020}

\usepackage[utf8]{inputenc} 
\usepackage[T1]{fontenc}    
\usepackage{hyperref}       
\usepackage{url}            
\usepackage{booktabs}       
\usepackage{amsfonts}       
\usepackage{nicefrac}       
\usepackage{microtype}      

\usepackage{amsmath}
\usepackage{graphicx}
\usepackage{siunitx}
\usepackage[numbers]{natbib}
\usepackage[small]{caption}
\usepackage[dvipsnames]{xcolor}

\usepackage{amsmath,amsfonts,bm}

\def\eqref#1{equation~\ref{#1}}

\def\1{\bm{1}}

\DeclareMathAlphabet{\mathsfit}{\encodingdefault}{\sfdefault}{m}{sl}
\SetMathAlphabet{\mathsfit}{bold}{\encodingdefault}{\sfdefault}{bx}{n}

\usepackage[inline]{enumitem}
\usepackage{wrapfig}

\title{Semi-supervised reward learning for \\ offline reinforcement learning}

\author{
Ksenia Konyushkova \\
\texttt{kksenia@google.com} \\
\And
Konrad \.Zo\l{}na 
\And
Yusuf Aytar 
\And
Alexander Novikov 
\AND
Scott Reed
\And
Serkan Cabi
\And
Nando de Freitas
\AND
\normalfont DeepMind
}

\begin{document}

\maketitle

\begin{abstract}
In offline reinforcement learning (RL) agents are trained using a logged dataset.
It appears to be the most natural route to attack real-life applications because in domains such as healthcare and robotics interactions with the environment are either expensive or unethical. 
Training agents usually requires reward functions, but unfortunately, rewards are seldom available in practice and their engineering is challenging and laborious. 
To overcome this, we investigate reward learning under the constraint of minimizing human reward annotations.
We consider two types of supervision: timestep annotations and demonstrations.
We propose semi-supervised learning algorithms that learn from limited annotations and incorporate unlabelled data.
In our experiments with a simulated robotic arm, we greatly improve upon behavioural cloning and closely approach the performance achieved with ground truth rewards. 
We further investigate the relationship between the quality of the reward model and the final policies.
We notice, for example, that the reward models do not need to be perfect to result in useful policies.
\end{abstract}
\section{Introduction}
\label{sec:intro}

\begin{wrapfigure}{t}{0.5\textwidth}
  \centering
  \vspace{-1.2cm}
  \includegraphics[width=0.63\linewidth]{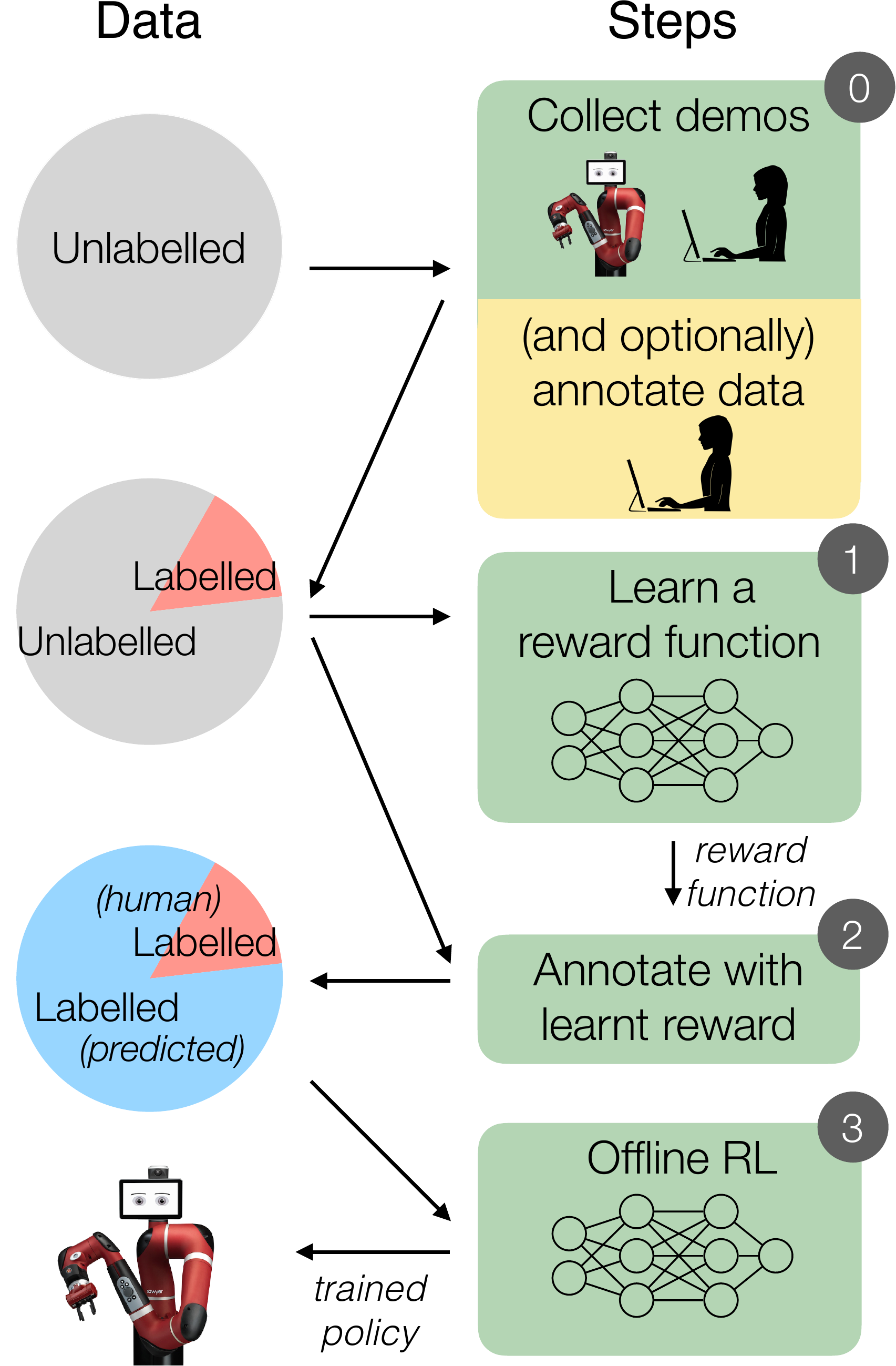}
  \caption{We train a reward function on a pre-recorded dataset, use it to label the data and do offline RL. As supervision, we use demonstrations, which can be optionally annotated with timestep rewards.}
  \label{fig:outline}
  \vspace{-.7cm}
\end{wrapfigure}

We are interested in applying deep reinforcement learning (RL) to obtain control policies from vision in realistic domains, such as robotic manipulation (\autoref{fig:mpg_tasks}). 
In online RL, agents are trained by interacting with an environment, while in offline RL, agents are trained using a logged dataset.
Our work builds upon offline RL~\cite{lange2012batch, levine2020offline, wang2020critic}, which seems to be well suited for real-life applications. 
However, there are some obstacles in applying offline RL to the real world: most algorithms require a reward signal which is often unavailable. 
To address this issue, we draw inspiration from the literature in inverse RL and reward modelling to learn a reward function. 
Then, it is used to retrospectively label episodes with rewards to make them available for offline RL (\autoref{fig:outline}). 
Both rewards and policies can be learned with the same logged dataset. 

We investigate various types of supervision for reward learning and how to use them efficiently.
In particular, we consider timestep-level (\emph{e.g.}, per-frame reward annotations for the entire episode produced by humans~\cite{cabi2020sketchy}, Sec.~\ref{sec:method_fromannotations}) or episode-level supervision (\emph{e.g.}, annotations of success for the whole episode, usually demonstrations~\cite{ORIL}, Sec.~\ref{sec:method_fromdemo}).
Timestep annotations are laborious to produce, but they contain a precise signal for learning.
Episode annotations are usually easier to obtain, but they provide only limited information about the reward. That is, they indicate that some of the video frames from the episodes show successful behaviour, but do not indicate when the success occurs. 
To deal with the setting of minimal reward supervision, we investigate sample efficient semi-supervised learning algorithms that benefit from large unlabelled datasets. 
Inspired by multiple-instance learning, co-training and self-supervision, we further propose an algorithm to deal with coarse reward labels in demonstrations, which uses time structure of the episodes and refines label predictions iteratively. 

We demonstrate that efficient manipulation policies can be obtained even when rewards are learnt with limited supervision. 
If a sufficient amount of episode-level annotations is available, timestep annotations are not needed to produce satisfactory policies. 
However, in a setting where the demonstrated trajectories are rare, good policies can still be obtained after incorporating more detailed timestep-level labels.
Further, we examine the reward models and their influence on the offline RL agent performance.
We inspect the various supervised learning metrics for reward models and their correlation with agent performance. 
Interestingly, we notice that our offline RL algorithm can be quite forgiving of the imperfections in reward models. 
\section{Related work}

\paragraph{Offline RL}
Offline RL enables learning policies from the logged data instead of collecting it online~\citep{lange2012batch, levine2020offline}. 
It is a promising approach for many real-world applications. 
Offline RL is an active area of research and many algorithms have been proposed recently, \emph{e.g.}, BCQ~\cite{fujimoto2019off}, MARWIL~\cite{wang2018exponentially}, BAIL~\cite{chen2019bail}, ABM~\cite{siegel2020keep} AWR~\cite{peng2019advantage}, and CRR~\cite{wang2020critic}.
In our work, we adopt CRR due to its efficiency and simplicity.
However, unlike standard offline RL, we do not observe the task rewards. 

\vspace{-2mm}
\paragraph{Reward learning}
When the reward signal is not readily available in the environment, it can be learnt.
If demonstrations are given, the reward can be learnt either directly with inverse RL~\citep{ng2000algorithms, abbeel2004apprenticeship}, or indirectly with generative adversarial imitation learning (GAIL)~\citep{ho2016generative}. 
If the end goal~\citep{edwards2016perceptual, singh2019endtoend} or reward values~\cite{cabi2020sketchy} are known for a subset of episodes, reward functions could be learnt with supervised learning.
Some works study the important case of learning from limited reward supervision~\cite{singh2019endtoend}.
Rewards are often learnt for the purposes of online RL.
While a lot of success has been achieved in learning from engineered or pre-trained state representations~\citep{edwards2016perceptual, finn2016guided, sermanet2017unsupervised, li2017infogail, fu2017learning, merel2017learning, zhu2018reinforcement, baram2017end}, learning directly from pixel input is known to be still challenging~\citep{zolna2019task} and the amount of required supervision may become a bottleneck~\cite{singh2019endtoend}.
Unlike many other reward learning approaches, we focus on learning from \emph{pixel} input in \emph{offline} RL from \emph{limited and coarse} annotations. 

\vspace{-2mm}
\paragraph{Behavioural cloning (BC)}
When demonstrations, but no reward signal is available, policies could be learnt with imitation learning~\citep{osa2018an}, and in particular, behavioural cloning (BC)~\citep{pomerleau1989alvinn, rahmatizadeh2018vision}.
BC is a supervised learning technique that aims to imitate the exact actions observed in demonstrations. 
One limitation of BC is that it requires large high-quality demonstration datasets that cover all the state space. 
In our experiments reward functions are learnt from the limited demonstrations and perform superior to BC.

\vspace{-2mm}
\paragraph{Learning from noisy labels}
When learning rewards from demonstrations, we learn from group labels, which are not as informative as direct timestep annotations. 
To efficiently utilise the available supervision, some ideas can be borrowed from self-training, multiple-instance learning and co-training. 
As in self-training~\cite{scudder1965probability, xie2020selftraining}, we use the predictions of a learnt function to guide its refinement.
Multiple-instance learning~\citep{kotzias2015fromgroup, cinbis2016mil} operates with an assumption that group labels are available and the task is to recover instance-level labels. 
Co-training~\citep{blum1998cotraining} usually divides features in several groups and uses predictions from one group as supervision for another group. 
PU-learning (positive unlabelled learning)~\citep{elkan2008learning, kiryo2017positive} and TRAIL (task relevant adversarial imitation learning)~\citep{zolna2019task} also deal with coarse training signal and they were successfully applied to adversarial imitation learning~\citep{xu2019positive, zolna2019task}.  

\vspace{-2mm}
\paragraph{Reward models with offline RL}
It is natural to learn reward functions in offline RL as both reward and policy training can rely on the same pre-recorded dataset (\autoref{fig:outline}). 
Similar to our approach, the work of~\citet{cabi2020sketchy} and ORIL~\citep{ORIL} learn reward functions and use them in offline RL. 
\citet{cabi2020sketchy} employ a reward sketching interface to elicit human preferences and use them as a signal for learning. 
In reward sketching, the annotator draws a curve where higher values correspond to higher rewards. 
As the values are not very precise in absolute terms, the ranking loss is used for training.
In this work, our timestep annotations are binary and it allows us to treat the reward prediction as a classification problem.
Furthermore, we focus on sample efficiency as we deal with limited human supervision. 
A recent method ORIL~\citep{ORIL} goes further and obtains reward functions both from labelled and unlabelled data at the same time as training an agent.
It relies on demonstrated trajectories, and employs ideas from PU-learning~\citep{xu2019positive} and TRAIL~\citep{zolna2019task} to correct for group labels. 
We analyse ORIL as one of the methods with episode-level supervision. 
Here we study both types of annotations from these works~\citep{cabi2020sketchy, ORIL}. 
Additionally, we study how to select the most promising reward model.

\section{Method}
\label{sec:method}

\subsection{Overview}
\label{sec:method_overview}

We take advantage of the Markov Decision Process (MDP) inductive bias. In particular, we observe trajectories of states and actions \emph{without} rewards $\tau = (s_1, a_1, \ldots , s_T, a_T)$, but we are given a small amount of information about the reward, as we discuss in detail in Secs.~\ref{sec:method_fromdemo} and~\ref{sec:method_fromdemo}.
As in offline RL, such trajectories are logged in a dataset $\mathcal{D}$.
In practice, $\mathcal{D}$ includes diverse trajectories produced for various tasks by scripted, random or learnt policies as well as human demonstrations~\citep{cabi2020sketchy}. 
Our workflow is illustrated in~\autoref{fig:outline} and it consists of the following steps:
\begin{enumerate}
    \item Infer a reward function $R$ from a small amount of supervision, either (1) episode-level annotations in the form of a set of successful episodes $\mathcal{D}_E$, usually demonstrations (Sec.~\ref{sec:method_fromdemo}), or (2) timestep-level annotations in the form of reward values on a subset of trajectories (Sec.~\ref{sec:method_fromannotations}).
    \item Use $R$ to retrospectively annotate all the trajectories in $\mathcal{D}$ with the reward values $\hat{r}_t=R(s_t)$.
    \item Use the trajectories with the predicted rewards $\hat{r}$ to do offline RL (Sec.~\ref{sec:method_policy}).
\end{enumerate}

\subsection{Reward learning from episode-level supervision}
\label{sec:method_fromdemo}

Suppose that as a form of reward supervision we are given a small set $\mathcal{D}_E$ of successful trajectories (\emph{e.g.}, expert demonstrations). 
For simplicity, we assume that a reward is binary and it indicates if the task is solved~\footnote{The described methods can be adapted to numerical rewards~\cite{cabi2020sketchy} as well.}.
We can view $\mathcal{D}_E$ as episode-level labels: they do not indicate the value of any timestep reward $r_t$, but they indicate that at least one of them is positive. 
The rest of the trajectories $\mathcal{D}_U = \mathcal{D} \setminus \mathcal{D}_E$ are unlabelled, and they include both success and failure episodes.
Several ways can be used to obtain reward values based on $\mathcal{D}_E$ and $\mathcal{D}_U$.

\paragraph{Reward without learning (SQIL)}
Similar to \textbf{SQIL} which is reported to be successful in online RL~\citep{reddy2020squil}, without reward learning, we set every timestep in expert demonstrations $\mathcal{D}_E$ to have reward $1$ and every timestep in unlabelled trajectories $\mathcal{D}_U$ to have reward $0$:
\begin{equation}
    \begin{aligned}
        \forall \tau \in \mathcal{D}_E, \forall t: \bar{r}_t = 1 ; \\
        \forall \tau \in \mathcal{D}_U, \forall t: \bar{r}_t = 0 .
    \end{aligned}
    \label{eq:sqil}
\end{equation}

\paragraph{Learn from flat reward (ORIL)}

Instead of using the reward values $\bar{r}_t$ from \autoref{eq:sqil} directly in RL, it was proposed to use them as a training signal to learn a reward function $R$~\cite{ORIL}. 
This idea is similar to training a discriminator in GAIL, applied to the offline RL setting. 
Then, the loss function is the cross-entropy that treats $\bar{r}_t$ as synthetic ground truth labels:
\begin{align}
    \mathcal{L}(\mathcal{D}) = \mathbb{E}_{s_t \sim \mathcal{D}_E} [-\log R(s_t)] + \mathbb{E}_{s'_t \sim \mathcal{D}_U} [-\log (1 - R(s'_t)],
    \label{eq:loss_flat}
\end{align}
where $s_t \sim \mathcal{D}$ denotes a state $s_t$ that is sampled from a trajectory $\tau$ that is sampled from $\mathcal{D}$. 
\textbf{ORIL} method~\cite{ORIL} additionally introduces several types of regularizations, such as TRAIL and PU-learning, to mitigate the problem of false negative trajectories present in $\mathcal{D}_U$ and false positive timesteps in $\mathcal{D}_E$ (originating from the flat reward assumption). 
We adapt \textbf{ORIL} in our work by separating training a reward model from training an agent.

\paragraph{Time-guided reward (TGR)}

Next, we propose a semi-supervised algorithm for learning the reward (\textbf{TGR}) and an algorithm to further iteratively refine its predictions (\textbf{TGR-i}). 
Instead of assuming a flat positive reward at the whole duration of the demonstration, we propose to make use of the \emph{time} structure of episodes.
We set all successful episodes to have reward zero in all timesteps until time $t_0$ and reward one after that moment.
As before, unlabelled trajectories are assigned label zero. 
Formally:
\begin{equation}
    \begin{aligned}
        \forall \tau \in \mathcal{D}_E, \forall t \le t_0 : \bar{r}_t = 0 ; \\
        \forall \tau \in \mathcal{D}_E, \forall t > t_0 : \bar{r}_t = 1 ; \\
        \forall \tau \in \mathcal{D}_U, \forall t : \bar{r}_t = 0 .
    \end{aligned}
    \label{eq:time_reward}
\end{equation}
Hyperparameter $t_0$ is shared among the episodes, its value is chosen according to the model selection procedure as described in Sec.~\ref{exp-reward}.
Then, based on the synthetic labels from~\autoref{eq:time_reward} the loss is:
\begin{equation}
    \begin{aligned}
        \mathcal{L}(\mathcal{D}) = \mathbb{E}_{s_t \sim \mathcal{D}_E: t \le t_0} [-\log (1-R(s_t))] +\mathbb{E}_{s'_t \sim \mathcal{D}_E: t > t_0} [-\log R(s'_t)] + \\ 
        \mathbb{E}_{s''_t \sim \mathcal{D}_U} [-\log (1 - R(s''_t)].
    \end{aligned}
    \label{eq:loss_time}
\end{equation}
Arguably, \textbf{TGR} is simpler than \textbf{ORIL} as it does not need any special type of regularisation.

\paragraph{Refining reward function (TGR-i)}

In \textbf{TGR} we made an assumption about the reward structure to produce synthetic labels (\autoref{eq:time_reward}).
We propose to further refine the reward function training by using the ideas from self-training, co-training and MIL~\citep{cinbis2016mil}. 
As in cross-validation~\citep{blum1998cotraining, cinbis2016mil}, we split the data into disjoint sets and train separate models that help to refine each other further.

We randomly split $\mathcal{D}_E$ and $\mathcal{D}_U$ into two parts each: $\mathcal{D}_E^A, \mathcal{D}_E^B$, and $\mathcal{D}_U^A, \mathcal{D}_U^B$. 
Then, we minimize the loss from \autoref{eq:loss_time} to obtain a reward classifier $R^A_0$ trained on the trajectories from $\mathcal{D}^A = \mathcal{D}_E^A \cup \mathcal{D}_U^A$ and $R^B_0$ trained on the trajectories from $\mathcal{D}^B = \mathcal{D}_E^B \cup \mathcal{D}_U^B$.
Next, we ``cross-apply'' the classifiers to the datasets on which they were not trained. 
It produces new reward estimates:
$\forall s_t \in \mathcal{D}^B: \hat{r}^0_t=R^A_0(s_t)$ and $\forall s_t \in \mathcal{D}^A: \hat{r}^0_t=R^B_0(s_t)$. 
Now, we use the predicted reward values $\hat{r}^0_t$ as new synthetic labels for the next generation of the classifiers $R_1^B$ and $R^A_1$. 
Splitting into disjoint sets $A$ and $B$ is important as it prevents each reward model from overfitting to synthetic labels on the data for which the predictions are used to train new models.
This procedure is repeated several times, at each iteration $i \ge 1$ using the loss function with the reward labels produced at the previous iteration:
\begin{align}
    \mathcal{L}_i(\mathcal{D}^A) = \mathbb{E}_{s_t \sim \mathcal{D}^A} [ - R^B_{i-1}(s_t) \log R^A_i(s_t) - (1-R^B_{i-1}(s_t)) \log (1 - R^A_i(s_t))].
    \label{eq:loss_mil}
\end{align}
\vspace{-6mm}
\begin{align}
    \mathcal{L}_i(\mathcal{D}) = \mathcal{L}_i(\mathcal{D}^A) + \mathcal{L}_i(\mathcal{D}^B).
    \label{eq:loss_mil_total}
\end{align}
The learning problem is then: $\min_{R^A_i, R^B_i} L_i(D)$, where $R^A_{i-1}$ and $R^B_{i-1}$ stay fixed.
We refer to the refined reward function as \textbf{TGR-i}, where $i$ indicates the iteration of refinement.

\subsection{Reward learning from timestep-level supervision}
\label{sec:method_fromannotations}

Now, in addition to the episode-level annotations, we allow for timestep-level annotations. 
That is, for any trajectory $\tau = (s_1, a_1, \ldots, s_T, a_T)$ from a selected subset $\mathcal{D}_0$ we provide $\tau_{\bar{r}} = (s_1, a_1, \bar{r}_1 \ldots, s_T, a_T, \bar{r}_T)$. 
In practice, the labels $\bar{r}$ can be obtained, for example, with reward sketching procedure~\cite{cabi2020sketchy} or by specifying the goal states~\cite{edwards2016perceptual}.
Then, we minimise supervised learning loss on the synthetic labels $\bar{r}_t$:
\begin{align}
    \mathcal{L}_{sup}(\mathcal{D}_0) = \mathbb{E}_{s_t \sim \mathcal{D}_0} [- \bar{r}_t \log R(s_t) - (1-\bar{r}_1) \log (1 - R(s_t)].
    \label{eq:loss_sup}
\end{align}

\paragraph{Supervised by demonstrations (sup-demo)}

In practice, when only a limited amount of supervision is possible, people often choose to first annotate demonstrated trajectories $\mathcal{D}_E$ as they exhibit the most informative behaviour~\cite{cabi2020sketchy}. 
Then, using the loss $\mathcal{L}_{sup}(\mathcal{D}_E)$ from \autoref{eq:loss_sup} we train a reward function \textbf{sup-demo}.

\paragraph{Semi-supervised by demonstrations (sup-and-flat)}

In addition to the limited reward labels in \textbf{sup-demo}, we propose to leverage the unlabelled data in the semi-supervised learning fashion, similar to how it is done with episode-level annotations. 
We set the flat zero synthetic reward for unlabelled subset $\forall s_t \sim \mathcal{D}_U: \bar{r}_t = 0$.
Then the loss is optimised jointly over timestep-level annotations in $\mathcal{D}_E$ and synthetic flat labels in $\mathcal{D}_U$, thus combining supervised and flat reward components (\textbf{sup-and-flat}):
\begin{align}
    \mathcal{L}(\mathcal{D}) = \mathbb{E}_{s_t \sim \mathcal{D}_E} [- \bar{r}_t \log R(s_t) - (1-\bar{r}_t) \log (1 - R(s_t)] + \mathbb{E}_{s'_t \sim \mathcal{D}_U} [-\log (1 - R(s'_t)].
    \label{eq:loss_semisup}
\end{align}
\subsection{Policy learning}
\label{sec:method_policy}

\paragraph{Critic-Regularised Regression (CRR)}
For learning a policy, we use a pre-recorded dataset $\mathcal{D}'$. 
$\mathcal{D}'$ contains some trajectories that were not seen during the reward training to avoid reward functions to overfit to the synthetic labels. 
We train an offline RL policy on the dataset with predicted rewards using Critic-Regularized Regression (CRR)~\cite{wang2020critic}, a state-of-the-art offline reinforcement learning method.
The policy update is a weighted version of BC, where the weights are determined by the learnt critic.

\paragraph{Behaviour cloning (BC)}
An alternative way to learn a policy when the reward values are not available is to use \textbf{BC}.
\textbf{BC} agent does not require reward values as it attempts to directly imitate the actions from the demonstrated trajectories $\mathcal{D}_E$.

\section{Experiments}
\label{exp}

In our experiments, we validate that
\begin{enumerate}
    \item Offline RL with learnt reward functions outperforms BC and RL with memorized rewards;
    \item Using unlabelled data is advantageous for reward learning with both types of reward supervision;
    \item The proposed \textbf{TGR-i} algorithm is among the top performing methods that utilise only episode-level annotations for training (Sec.~\ref{exp-policy-episode}) and it is quite robust to the choice of hyperparameters (Sec.~\ref{exp-reward});
    \item \textbf{TGR-i} with a sufficient amount of demonstrations matches the performance of the semi-supervised algorithm \textbf{sup-and-flat} that uses timestep reward annotations (Sec.~\ref{exp-policy-timestep});
    \item Offline RL policy training is forgiving of the reward model imperfections, but supervised learning metrics still help to guide the reward model selection (Sec.~\ref{exp-reward}).
\end{enumerate}

\subsection{Experimental setup}
\label{exp-setup}

\paragraph{Environment and tasks}

\begin{figure}
\centering
  \includegraphics[width=0.85\linewidth]{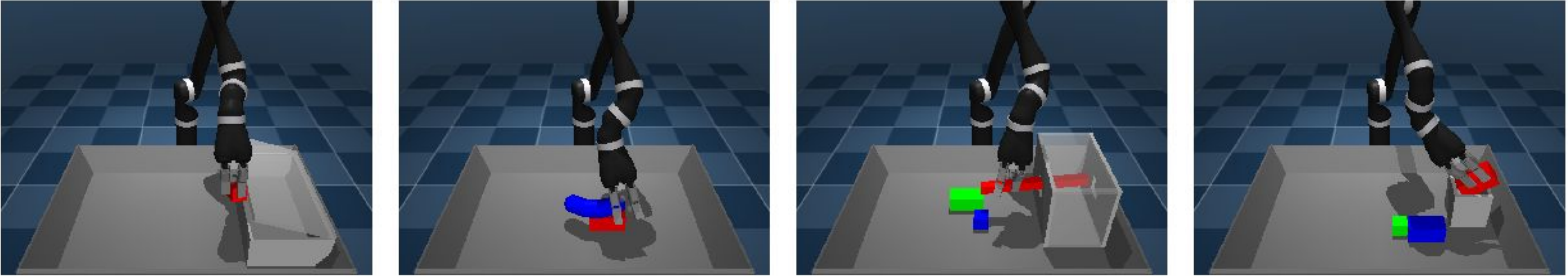}
  \caption{Kinova Jaco arm performs object manipulation: put into box, stack banana, slide, insert.}
  \label{fig:mpg_tasks}
  \vspace{-0.1cm}
\end{figure}

We conduct the experiments with a simulated Kinova Jaco arm with $9$ degrees of freedom.
We use joint velocity control of $6$ arm and $3$ hand joints. 
Note that we are interested in learning directly from pixel input and thus the agent needs to infer objects configuration from its camera observations.
There are two cameras: frontal and in-hand, each producing $64 \times 64$ pixel images.
Besides, we use the proprioception information.
We perform several manipulation tasks in $20 \times 20$ cm basket: \emph{put into box}, \emph{insertion}, \emph{slide}, and \emph{stack banana} (\autoref{fig:mpg_tasks}).
The environment reward is binary: $1$ if the task is solved and $0$ otherwise.

\paragraph{Datasets} 

The datasets are pre-generated according to a procedure from the work of~\citet{wang2020critic}.
Each episode terminates either \num{20} timesteps after the task is solved or after \num{400} timesteps.
The size of each dataset $\mathcal{D}'$ is around \num{8000} epsiodes.
Half of these episodes are available for reward learning, which includes demonstrations and unlabelled episodes.
The ground truth rewards are used to produce limited annotations, which are not observed for any other purpose. 
For simplicity, we define a successful trajectory to be such that at least one timestep has a positive reward. 
The demonstrations set $\mathcal{D}_E$ is selected among successful episodes where trajectories are sampled with probability $\frac{1}{16}$ for methods with episode-level annotations (Sec.~\ref{sec:method_fromdemo}) and with probability $\frac{1}{128}$ for methods with timestep-level annotations (Sec.~\ref{sec:method_fromannotations})~\footnote{Supplementary materials contain experiments with varying amounts of annotations.}. 
As a result of this sampling procedure, we have different amount of annotated trajectories in $4$ tasks. 
In total, we use \num{181}, \num{199}, \num{206} and \num{371} episode-level annotated episodes and \num{28}, \num{16}, \num{19} and \num{47} timestep-level annotated episodes for tasks \emph{box}, \emph{insertion}, \emph{slide}, and \emph{stack banana} respectively.

\paragraph{Baselines}

We summarize the methods in~\autoref{tab-baselines}. 
They are organised in the order of increasing amount of reward supervision. 
\textbf{GT} indicates an upper bound on the agent performance.
The other strategies are described in Sec.~\ref{sec:method}.

\begin{table}
  \caption{Summary of the methods used in our experiments. The methods are organised in the order of increasing amount of required reward supervision: a set of successful episodes, a set of successful episode with timestep annotations, ground truth reward signal.}
  \label{sample-table}
  \centering
  \begin{tabular}{lll}
    \toprule
    Name & Reward supervision & Description \\
    \midrule
    \textbf{BC} & A set of demonstrations  $\mathcal{D}_E$ & BC on expert demonstrations.  \\
    \textbf{SQIL} & A set of demonstrations $\mathcal{D}_E$ & CRR + memorised reward signal~\citep{reddy2020squil}. \\
    \textbf{ORIL} & A set of demonstrations $\mathcal{D}_E$ & CRR + flat reward + best regularisation~\cite{ORIL}. \\
    \textbf{TGR} & A set of demonstrations $\mathcal{D}_E$ & CRR + time-guided reward function. \\
    \textbf{TGR-i} & A set of demonstrations $\mathcal{D}_E$ & CRR + TGR model refined at $i$th iteration. \\  
    \midrule
    \textbf{sup-demo} & Timestep annotations on $\mathcal{D}_E$ & CRR + supervised reward model. \\
    \textbf{sup-and-flat} & Timestep annotations on $\mathcal{D}_E$ & CRR + semi-supervised reward model. \\
    \midrule
    \textbf{GT} & Ground truth rewards & CRR + ground truth reward \\
    \bottomrule
  \end{tabular}
  \label{tab-baselines}
\end{table}

\paragraph{Agent training}
All agents are trained with CRR algorithm~\citep{wang2020critic} (except for the BC agent) in Acme framework~\citep{hoffman2020acme}. 
The batch size is set to \num{1024} and the learning rate is \num{1e-4}. 
In every experiment we train a policy with \num{3} random seeds and show the rolling average score over \num{30}\% of data with the shading indicating where \num{95}\% of the datapoints lie.
\subsection{Learning from episode-level annotations}
\label{exp-policy-episode}

\begin{figure}
\centering
  \includegraphics[height=0.14\textheight]{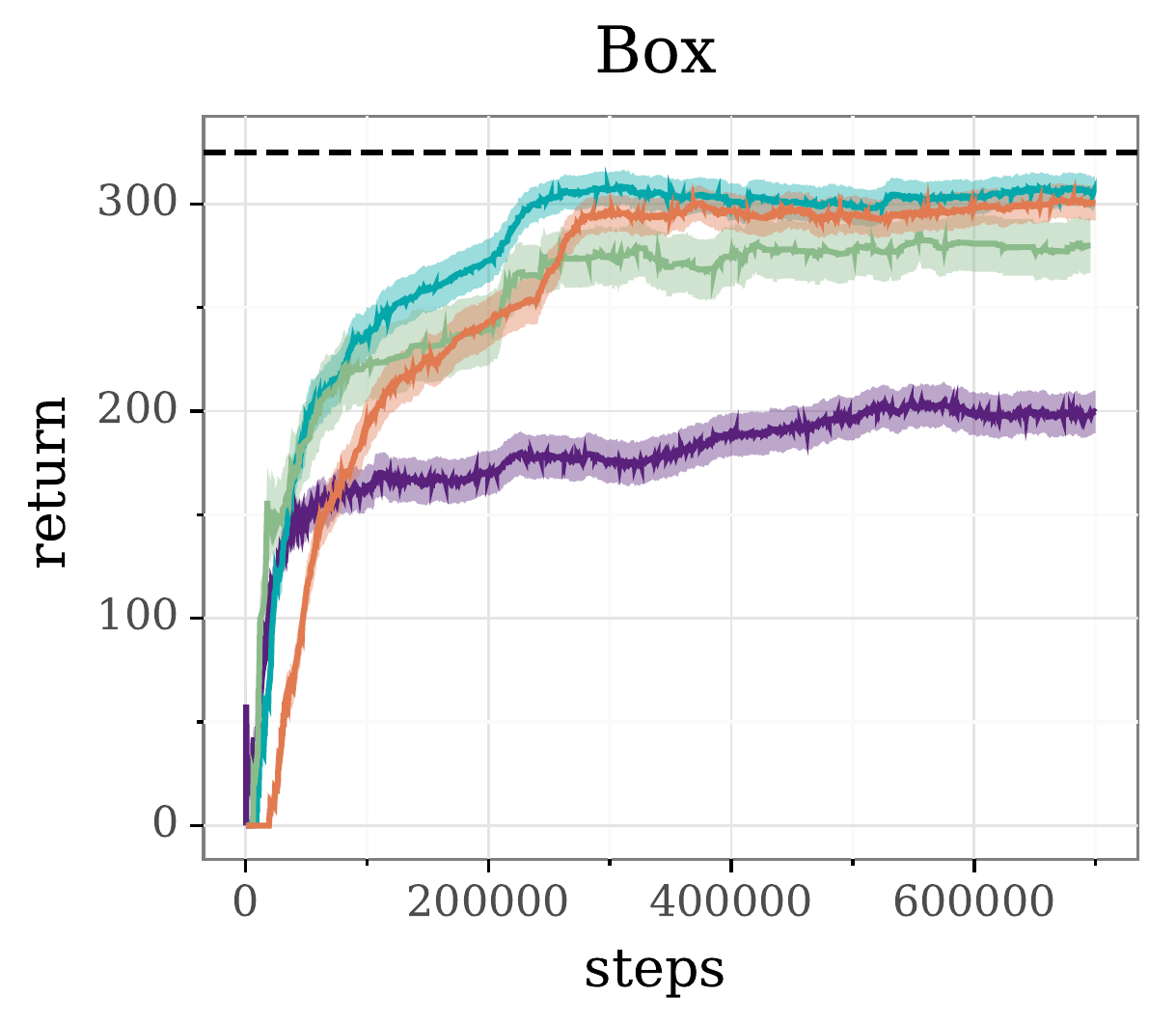}
  \includegraphics[height=0.14\textheight]{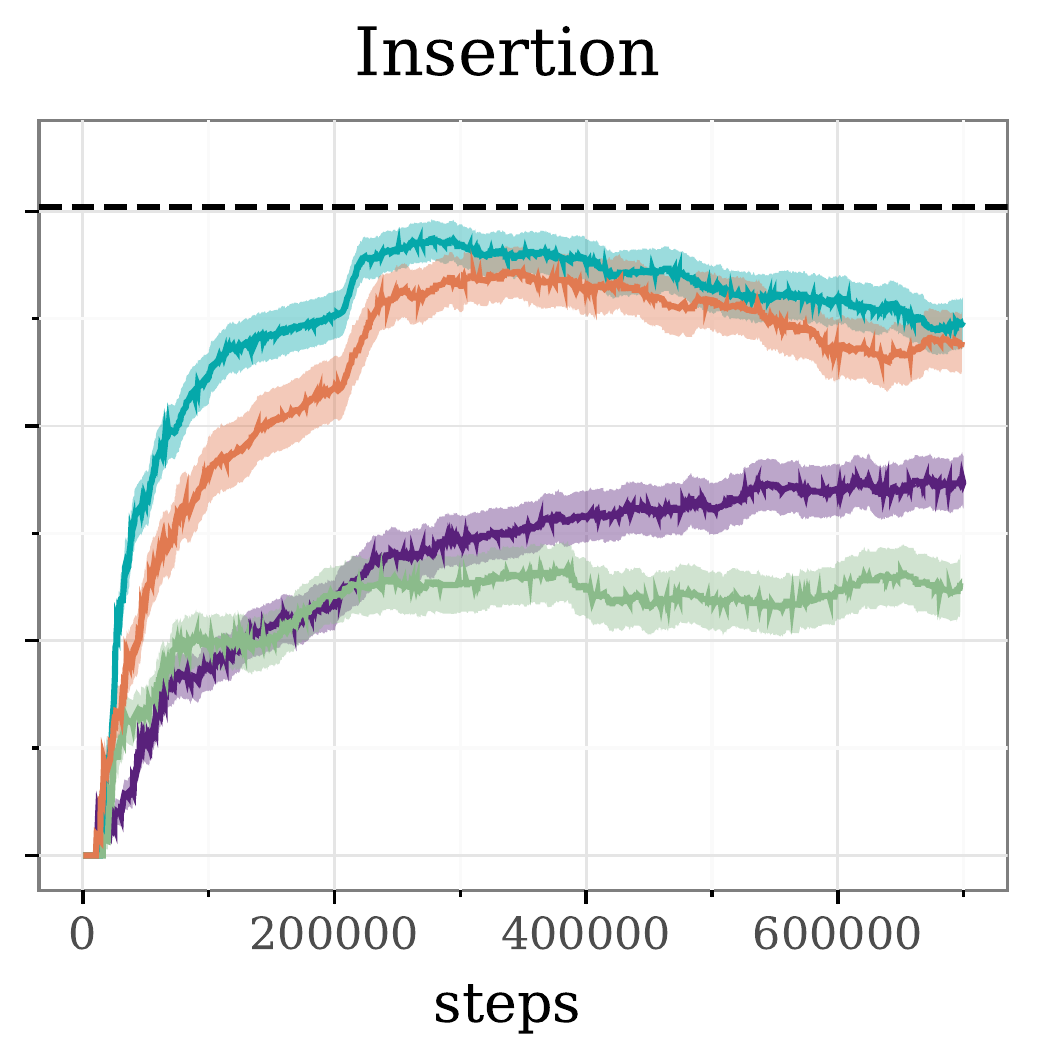}
  \includegraphics[height=0.14\textheight]{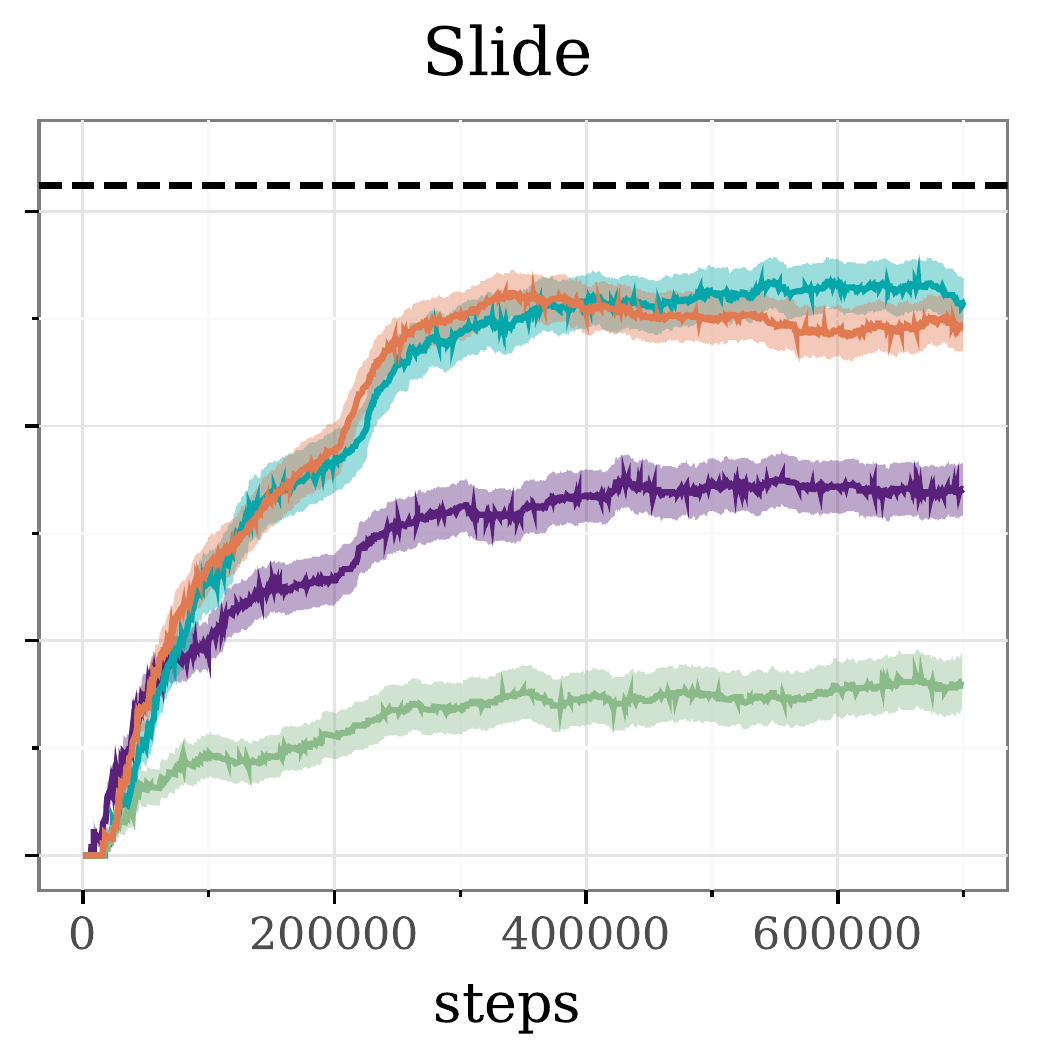}
  \includegraphics[height=0.14\textheight]{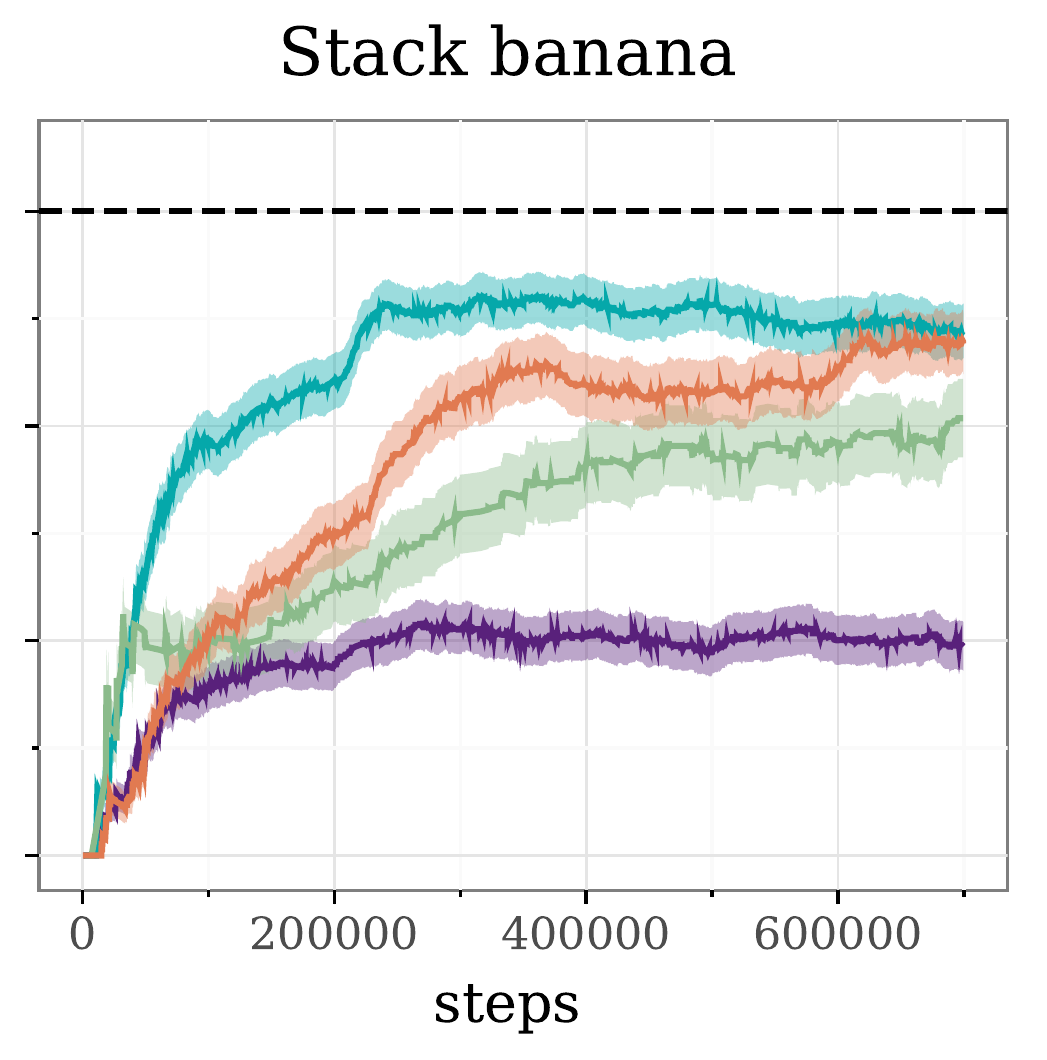}
  \includegraphics[width=0.4\linewidth]{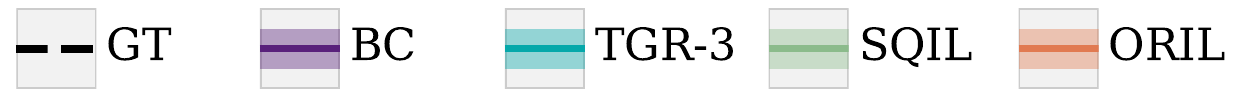}
  \caption{Policy training results for \num{4} tasks using different reward models trained with \emph{episode}-level labels. \textbf{TGR-i} strategy is consistently among the top performing models. 
  In addition to this, in several cases it closely approaches the performance achieved with ground truth labels.}
  \label{fig:results-group}
\end{figure}

In \autoref{fig:results-group} we show the performance of the policies that use episode-level annotations for determining rewards as described in Sec.~\ref{sec:method_fromdemo} and \autoref{tab-baselines}.
\textbf{BC} policy reaches only between $1/3$ and $2/3$ of the score attained by a CRR policy with ground truth rewards (\textbf{GT}). 
\textbf{SQIL} with memorized reward values works well in some cases (\emph{box}), but its performance is worse than \textbf{BC} in other cases (\emph{insertion}, \emph{slide}).
Learning a reward function from flat reward assumption with regularisation as in \textbf{ORIL} improves significantly over \textbf{BC}. 
Finally, \textbf{TGR-i} further improves the results making it to be among the best performing strategies across all tasks.
To sum up, semi-supervised reward learning techniques greatly reduce the gap to \textbf{GT} compared to \textbf{BC} while using exactly the same annotations.

\subsection{Learning from timestep-level annotations}
\label{exp-policy-timestep}

\begin{figure}
\centering
  \includegraphics[height=0.14\textheight]{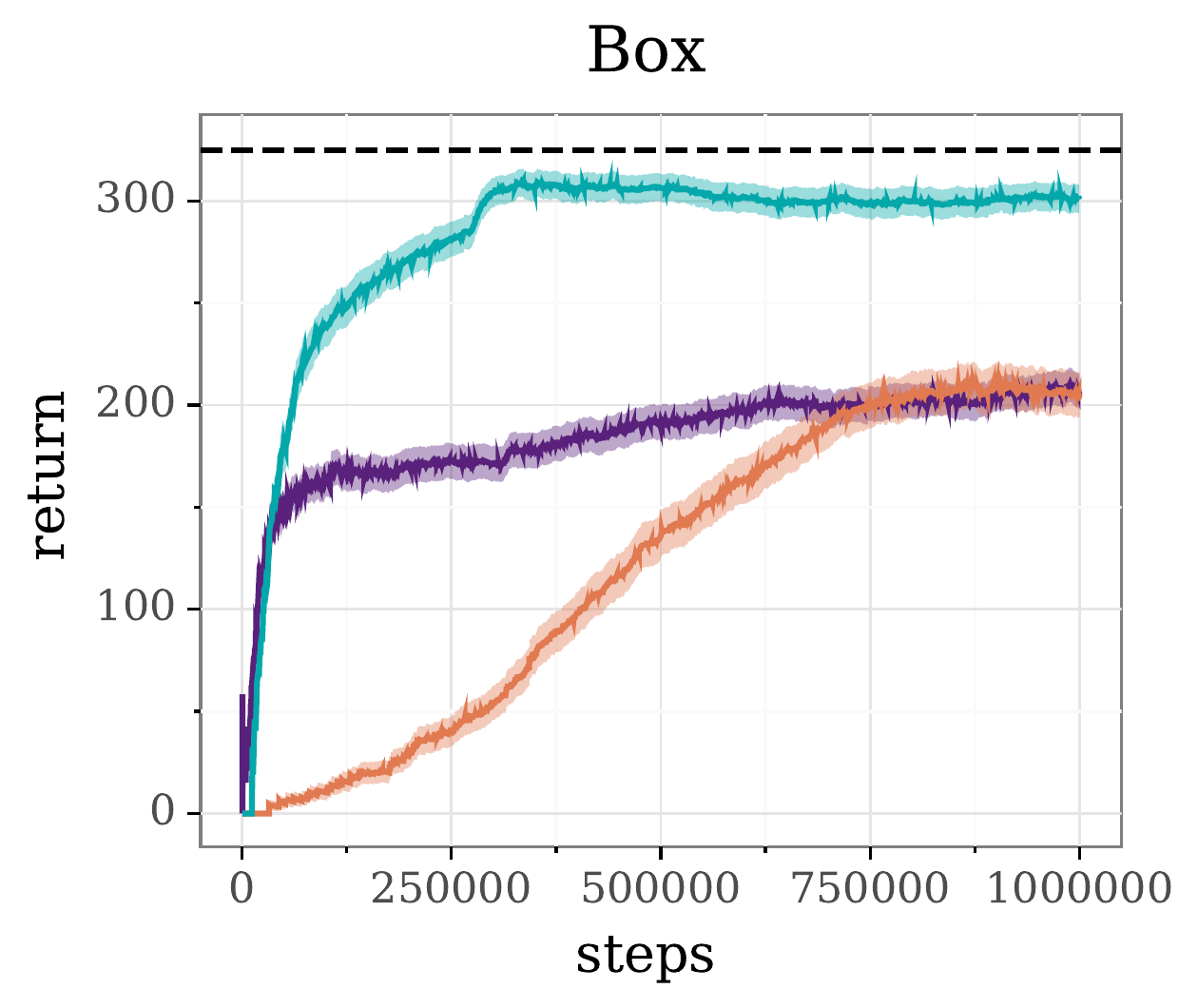}
  \includegraphics[height=0.14\textheight]{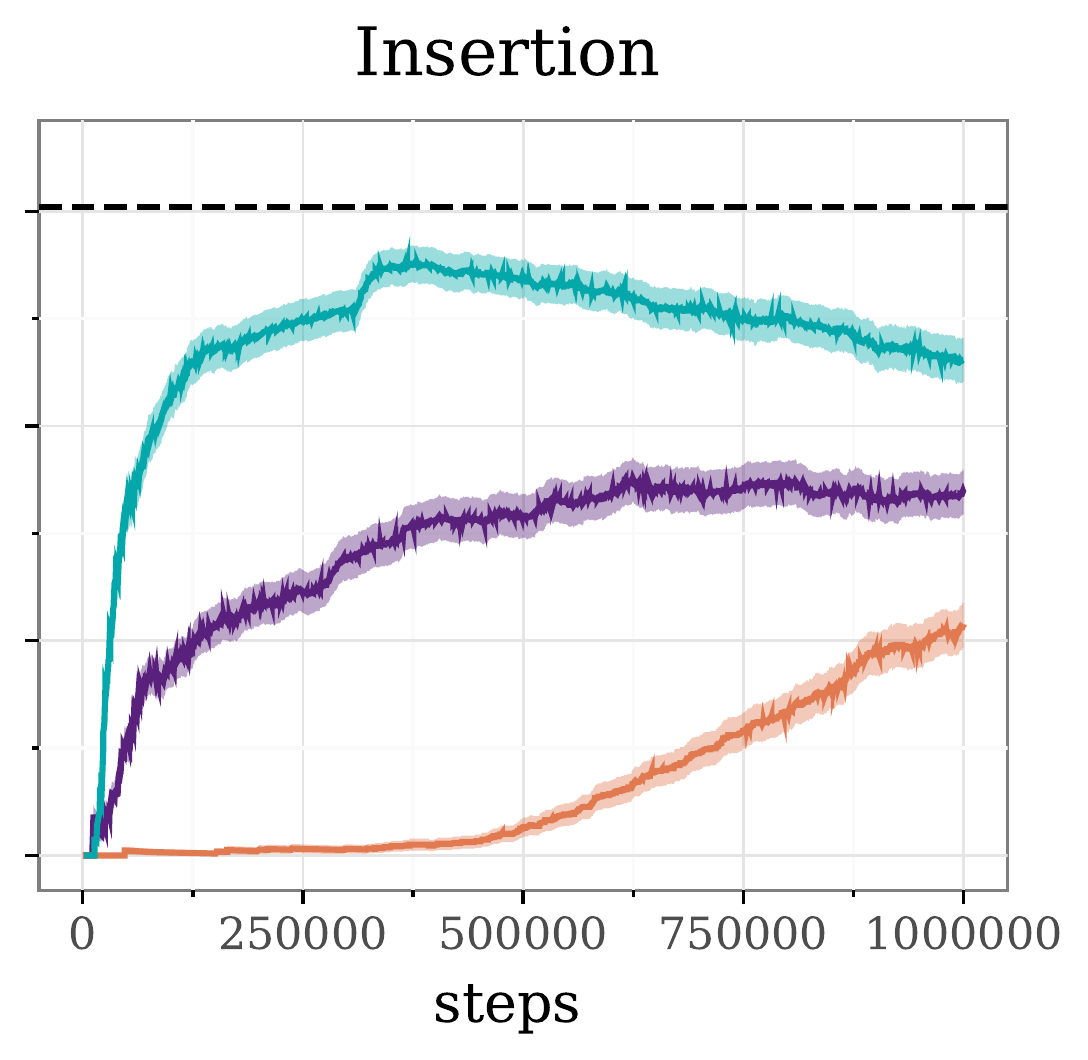}
  \includegraphics[height=0.14\textheight]{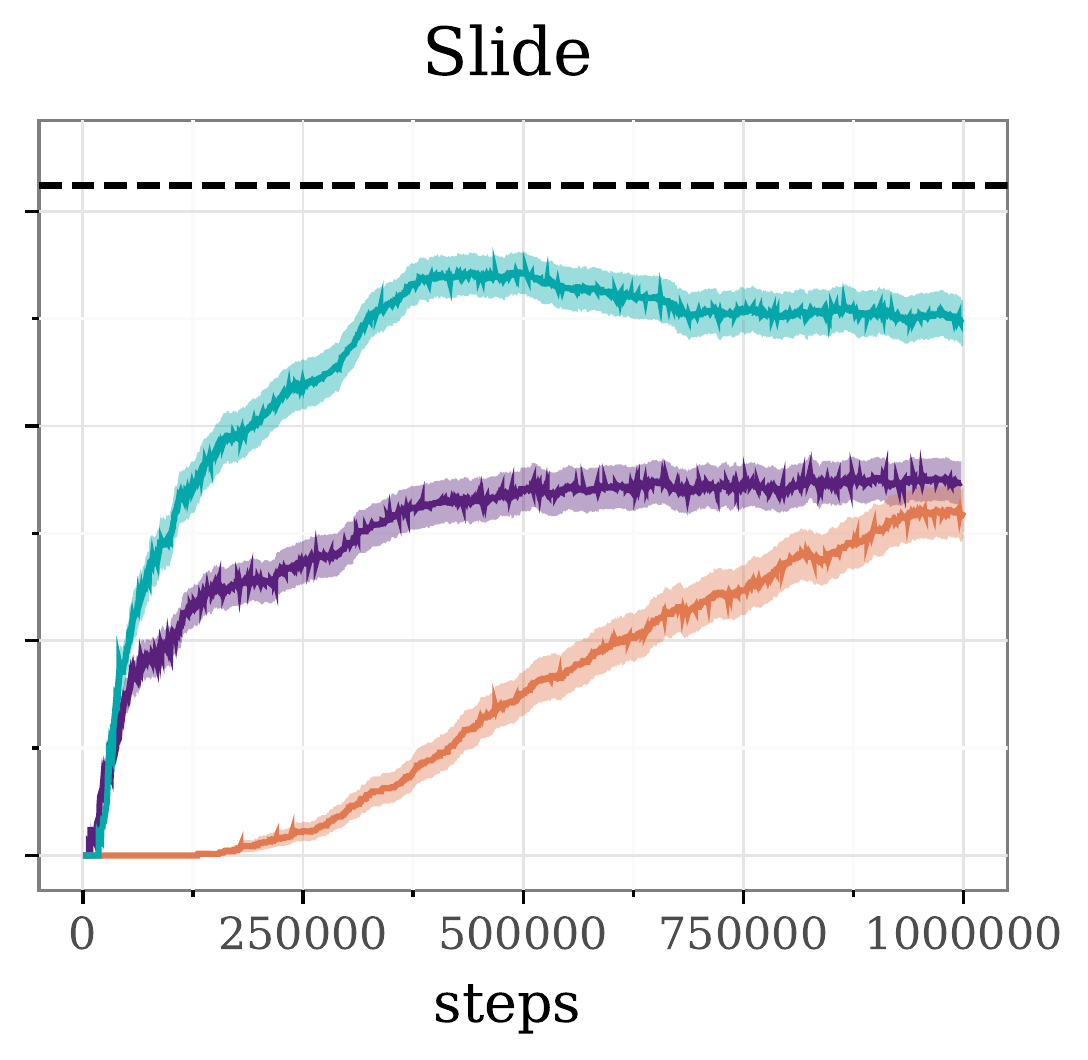}
  \includegraphics[height=0.14\textheight]{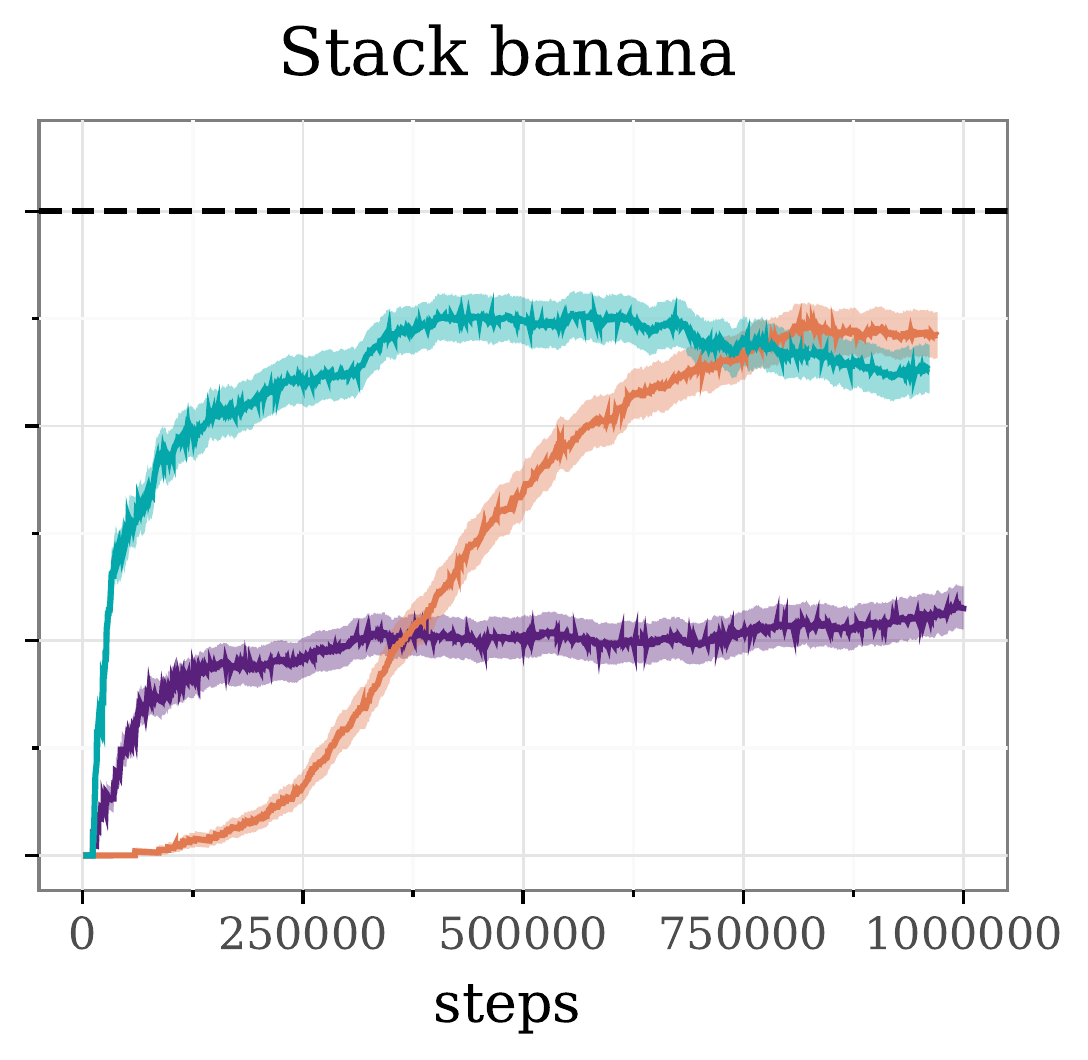}
 \includegraphics[width=0.45\linewidth]{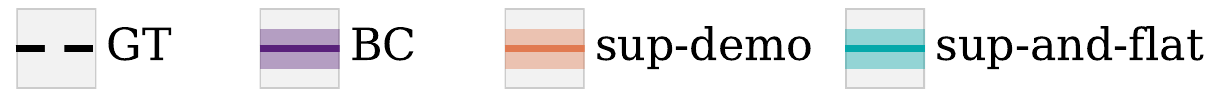}
  \caption{Results of policy training for \num{4} tasks using different reward model trained with \emph{timestep}-level labels. Leveraging the unlabelled data with small amount of annotation in \textbf{sup-and-flat} allows to attain high scores that closely approach the performance achieved with ground truth labels. }
  \label{fig:results-instance}
\end{figure}

\autoref{fig:results-instance} shows the performance of the policies that learn reward functions from timestep-level annotations described in Sec.~\ref{sec:method_fromannotations} and \autoref{tab-baselines}: \textbf{sup-demo} and \textbf{sup-and-flat}.
As before, for the reference, we indicate the performance of \textbf{GT} and \textbf{BC}. 
Both \textbf{sup-demo} and \textbf{sup-and-flat} use timestep annotations for \num{28}, \num{16}, \num{19} and \num{47} episodes for \emph{box}, \emph{insertion}, \emph{slide} and \emph{stack banana} tasks, but their performance differs greatly.
Using limited supervision in \textbf{sup-demo} sometimes does and sometimes does not match the scores of \textbf{BC} and only once performs on par with \textbf{sup-and-demo}.
Leveraging unlabelled data as in \textbf{sup-and-demo} brings a remarkable benefit to the agent's performance and allows it to closely approach \textbf{GT}. 

If we compare the results from Figs.~\ref{fig:results-instance} and~\ref{fig:results-group}, we notice that both \textbf{TGR-i} and \textbf{sup-and-flat} attain high policy scores while using the different types of supervision. 
In practice, it means that if a sufficient number (\num{200}--\num{300}) of demonstrations of a desired behaviour is available, there is no need to provide timestep annotations. 
However, if only a handful (\num{20}--\num{50}) of demonstrations is available, detailed annotations could help to match the performance.

\subsection{Reward model results}
\label{exp-reward}

Reward model quality can be assessed as the quality in the binary classification task of distinguishing between successful and failure timesteps. 
Thus, we track the classification quality metrics.
In the initial trials with \emph{accuracy} and \emph{recall} they turned out to be unsuccessful due to the imbalance of the classes. 
Using the \emph{validation loss} is biased because of the nature of synthetic training labels.
Thus, we compute:
\begin{enumerate*}[label=(\arabic*)]
    \item \emph{precision} as the proportion of the correctly classified timesteps among positive predictions, 
    \item \emph{f-score} as the combination of precision and recall, and
    \item \emph{AUC-PR} as the area under precision recall curve.
\end{enumerate*}
To estimate the scores, we sample and annotate a limited validation set from the unlabelled data: \num{89}, \num{101}, \num{76}, and \num{59} episodes for \emph{box}, \emph{insertion}, \emph{slide}, and \emph{stack banana} tasks. 

\paragraph{Model selection}

\begin{figure}
\centering
  \includegraphics[width=0.24\linewidth]{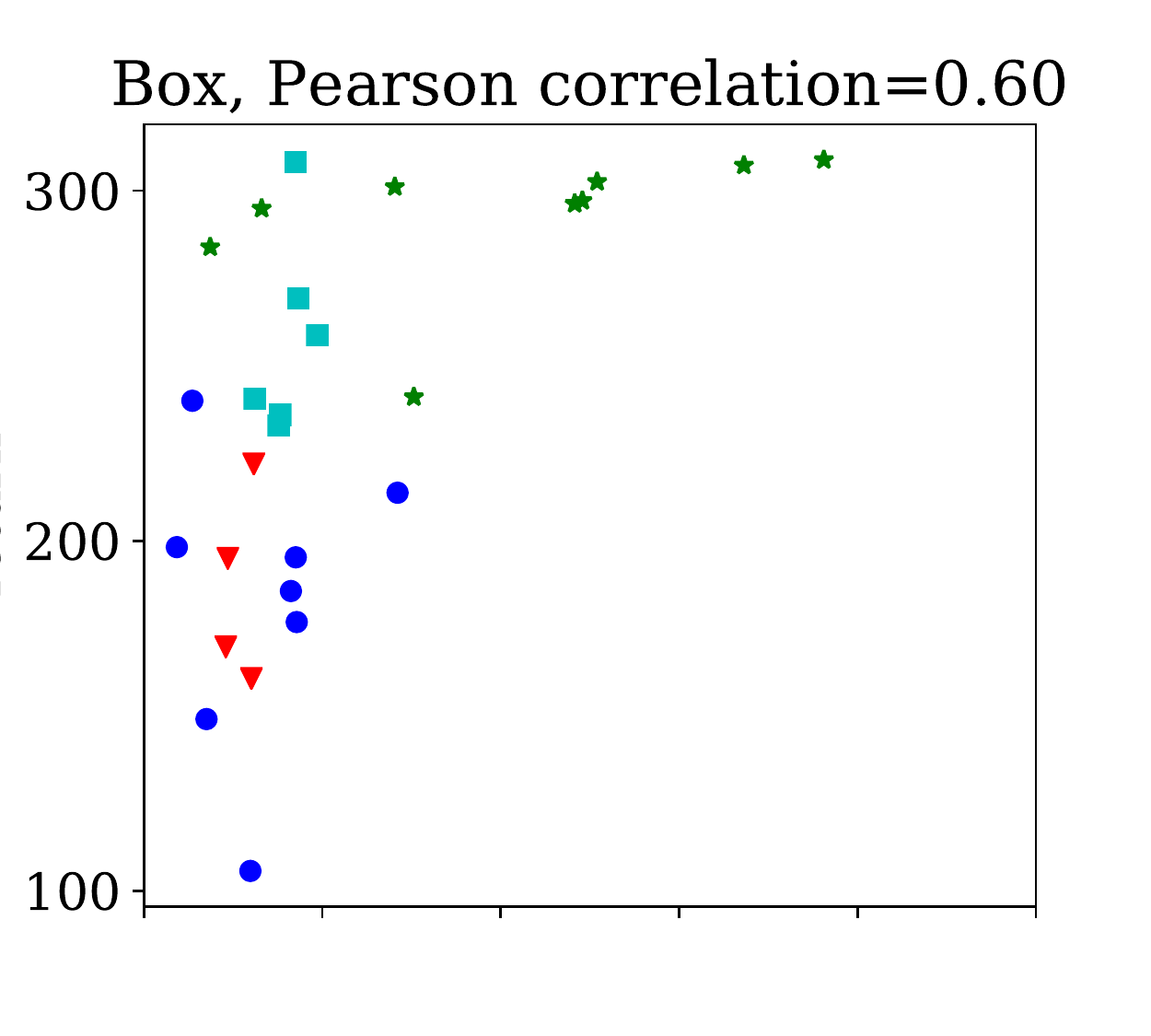}
  \includegraphics[width=0.24\linewidth]{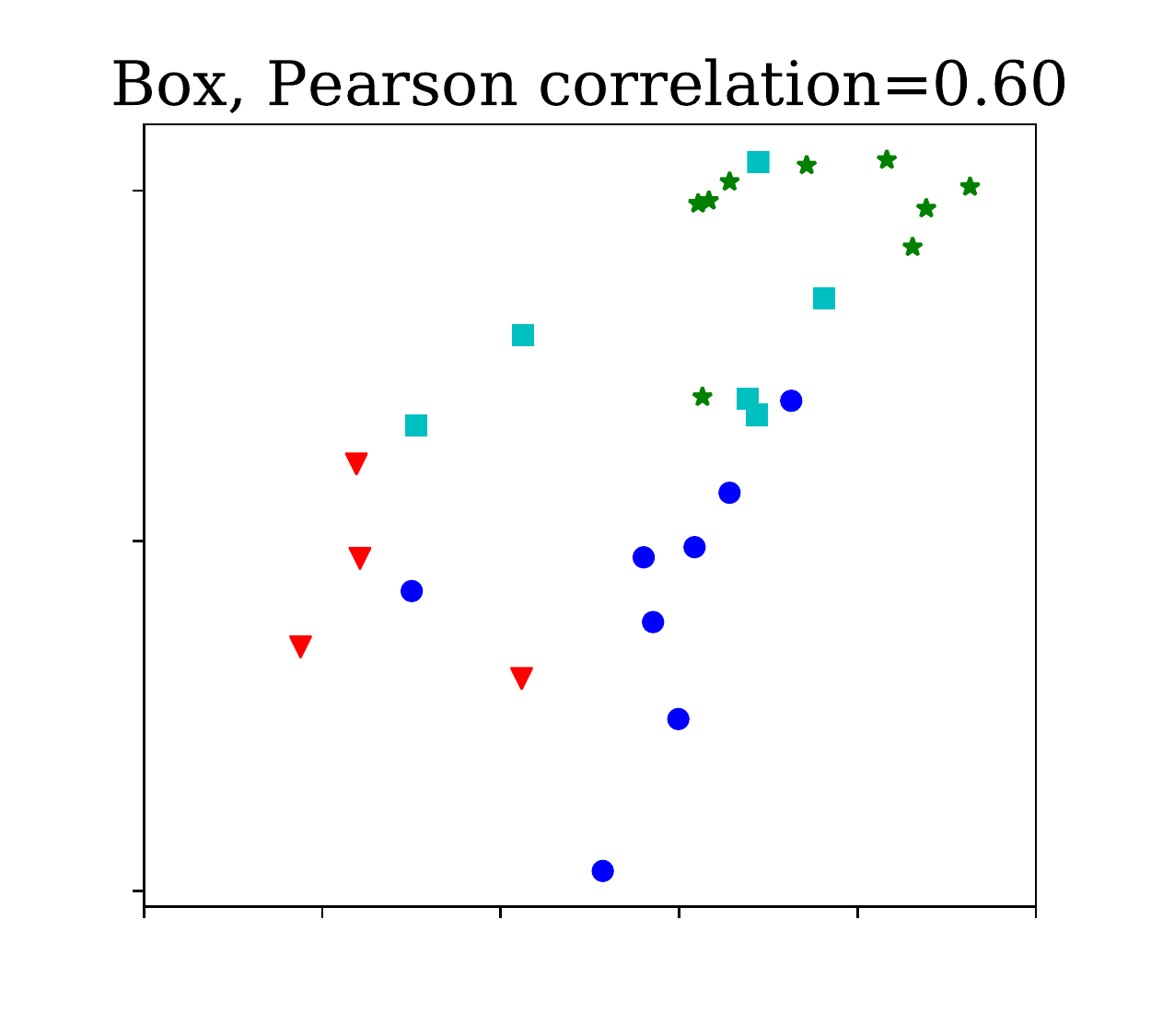}
  \includegraphics[width=0.24\linewidth]{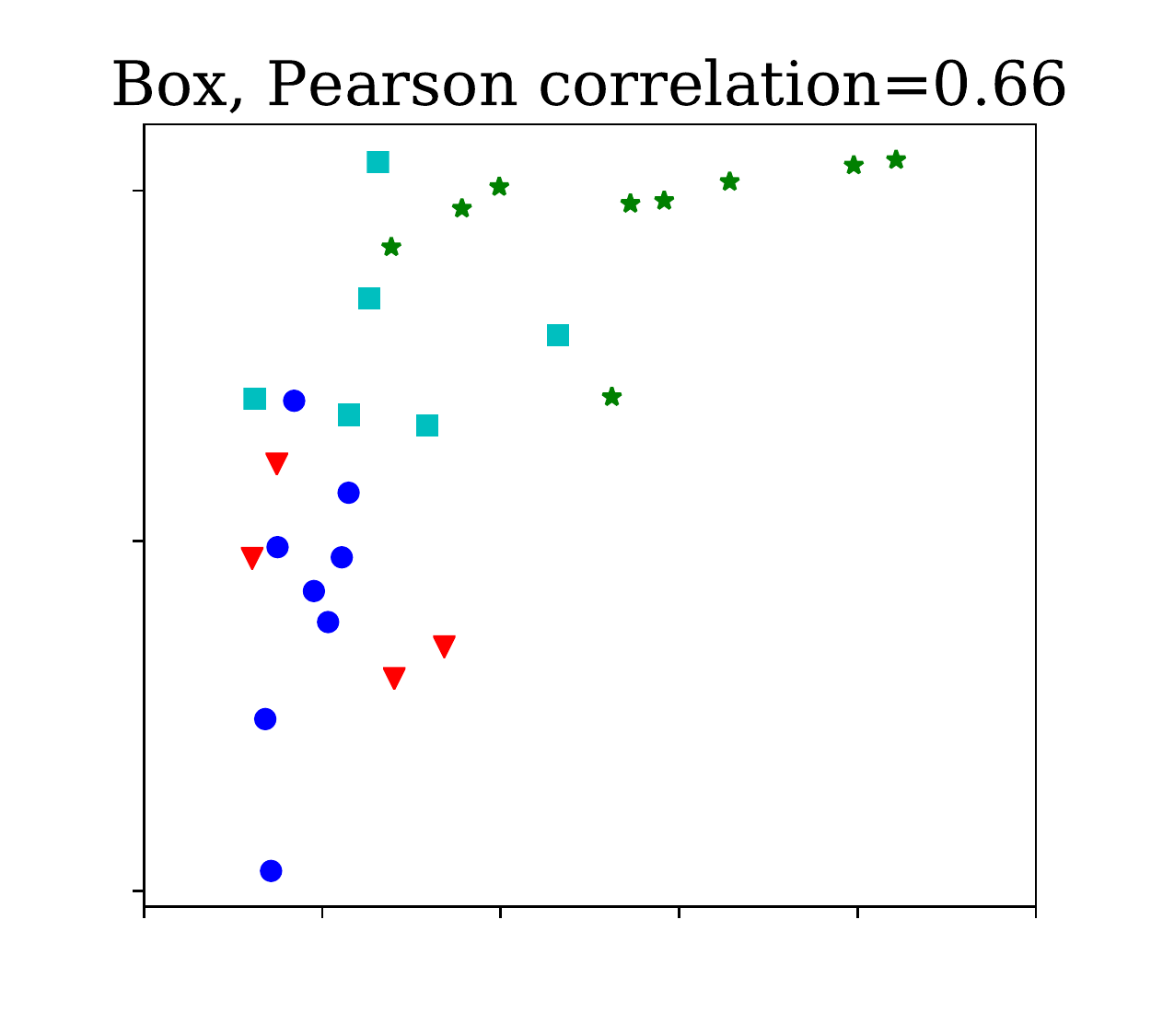} \\
  \vspace{-3mm}
  \includegraphics[width=0.24\linewidth]{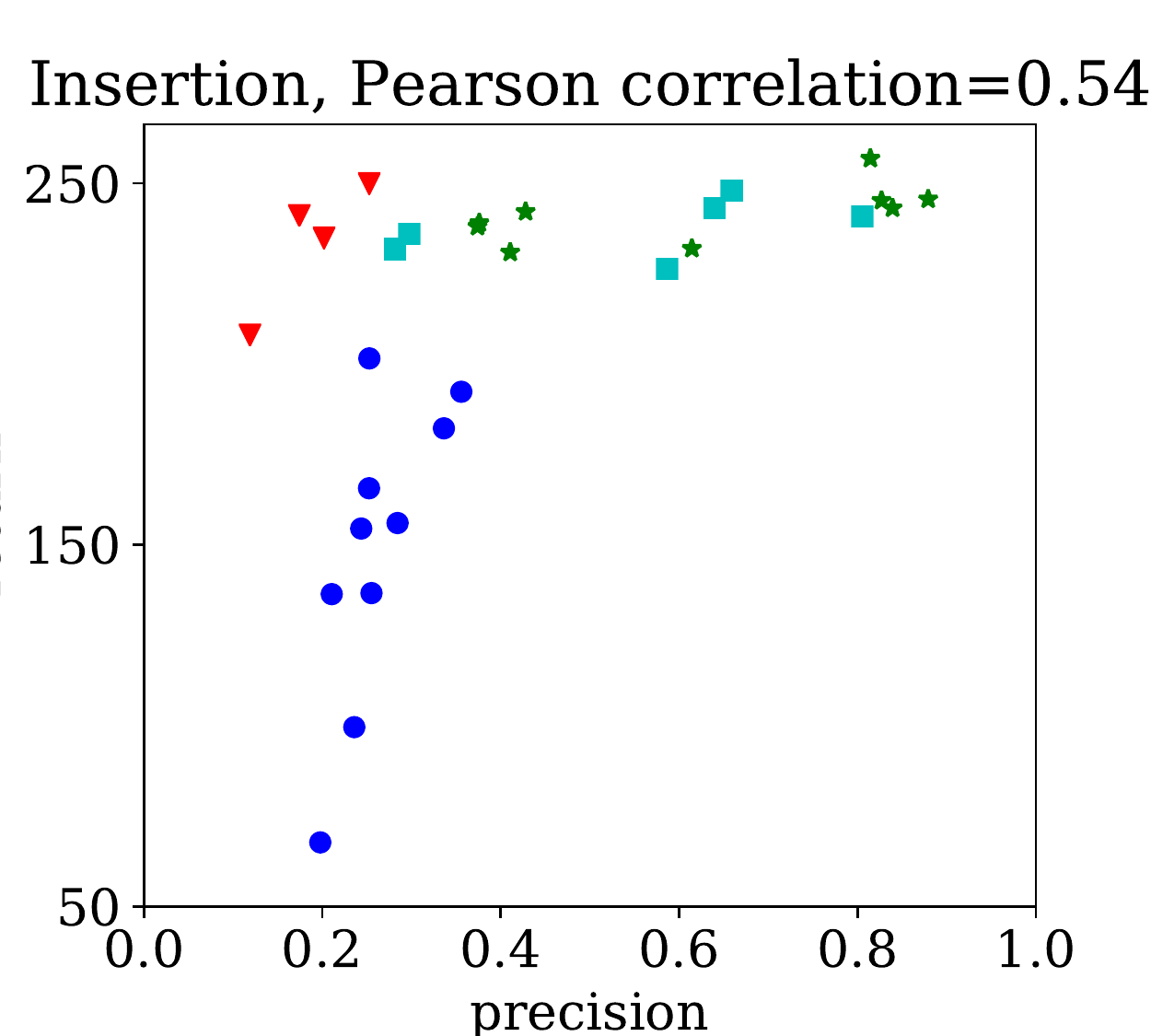}
  \includegraphics[width=0.24\linewidth]{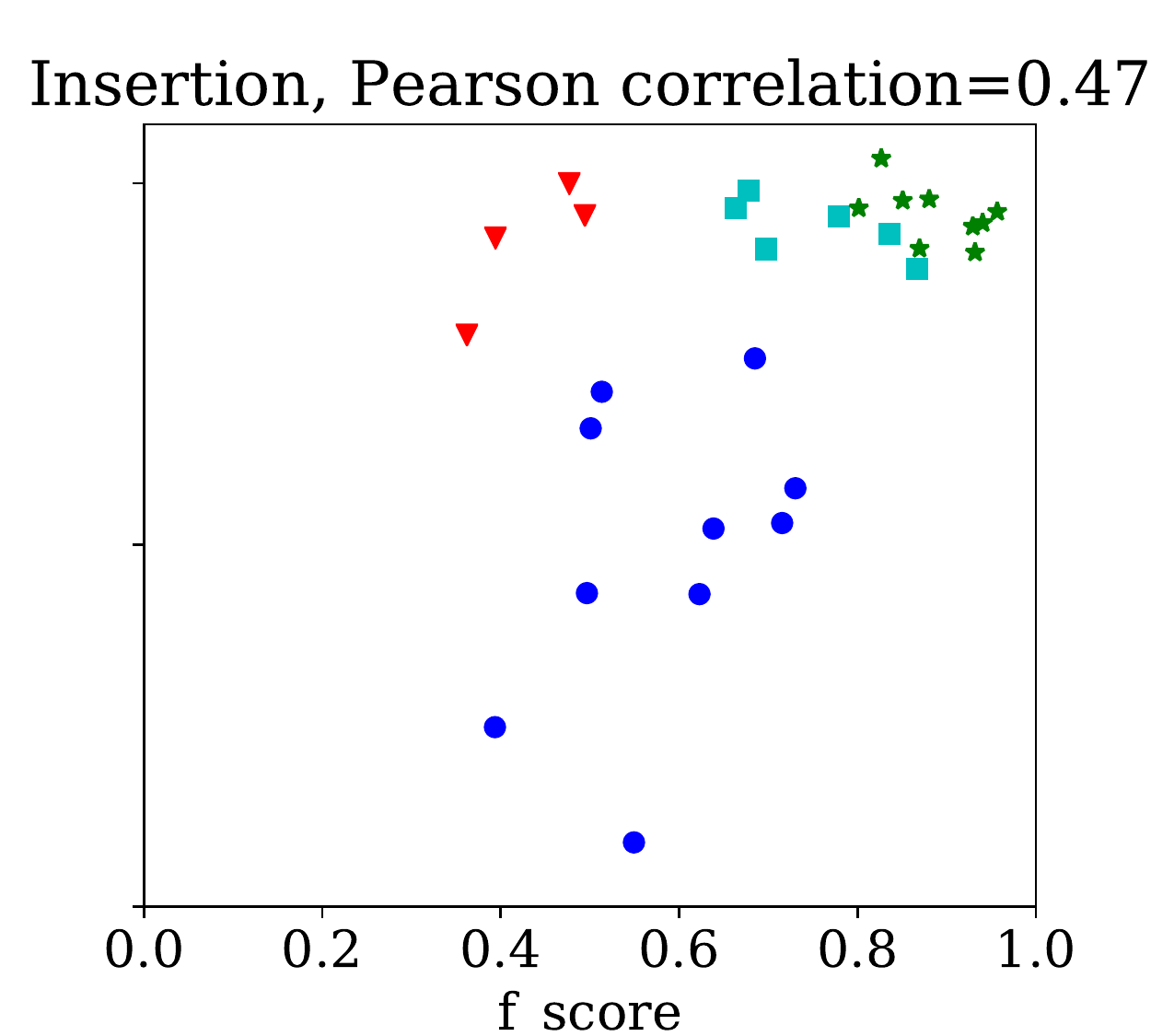}
  \includegraphics[width=0.24\linewidth]{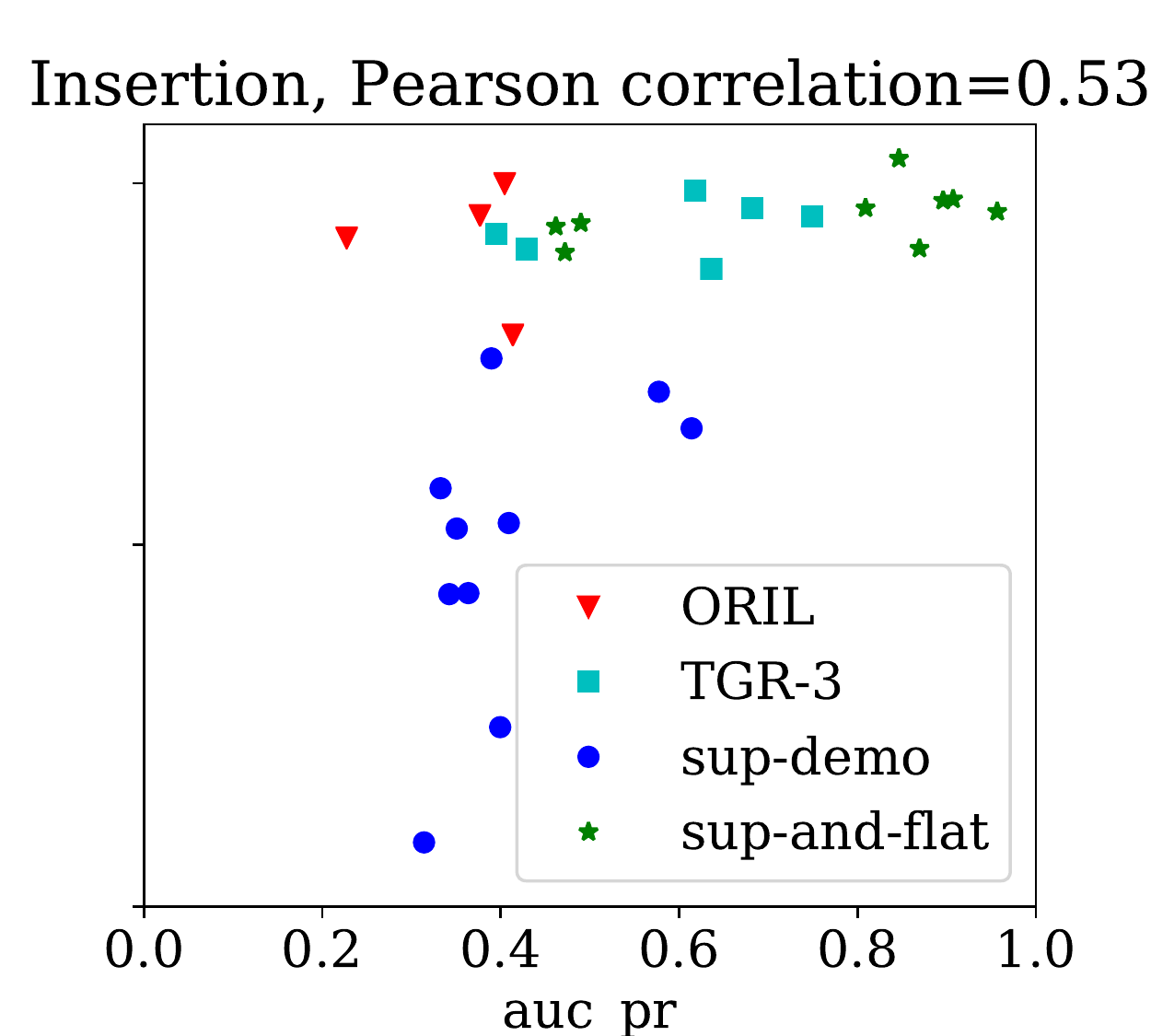}
  \caption{The relationship between the quality of the reward model (\emph{precision}, \emph{f-score}, and \emph{AUC-PR} in first, second, and third columns) and the performance of the policy in \textbf{box} (first row) and \textbf{insertion} tasks (second row). We take advantage of the correlation between the scores for the model selection.}
  \label{fig:reward-quality}
\end{figure}

Among the different hyper-parameter values, such as the type of regularization in \textbf{ORIL}, $t_0$ in \textbf{TGR}, learning rate and batch size, we select the classification models that achieve the best validation scores in each metrics.
The policy curves in Figs.~\ref{fig:results-group} and~\ref{fig:results-instance} show the best of reward models that maximize each of the metrics.
Then, we study how the quality of the reward models relates to the quality of the trained policies.
\autoref{fig:reward-quality} shows the dependency between these scores and the policy return at the end of the training (averaged of the last \num{10000} iterations) as well as their Pearson correlation.
The reward model scores are correlated with the agent performance, however, it is hard to identify which one would perform the best in advance.
Nevertheless, very high classification scores reliably indicate good policy performance (\emph{e.g.}, \textbf{sup-and-flat}).
The reward model scores of \textbf{ORIL} and \textbf{sup-demo} have similar spread, \textbf{TGR-i} scores are often higher and \textbf{sup-and-flat} method has the highest scores. 
Higher reward model scores (even without improvement in the policy) might be useful for different purposes, for example, for policy evaluation.  
Interestingly, sometimes reward models with low scores can still provide informative signal for RL training (\emph{e.g.}, \emph{precision}, \emph{f-score} and \emph{AUC-PR} of \textbf{ORIL} methods in \emph{insertion} task), thus, offline RL is forgiving of the noisy rewards. 

\paragraph{Robustness of iterative refinement}

\begin{figure}
\centering
  \includegraphics[height=0.145\textheight]{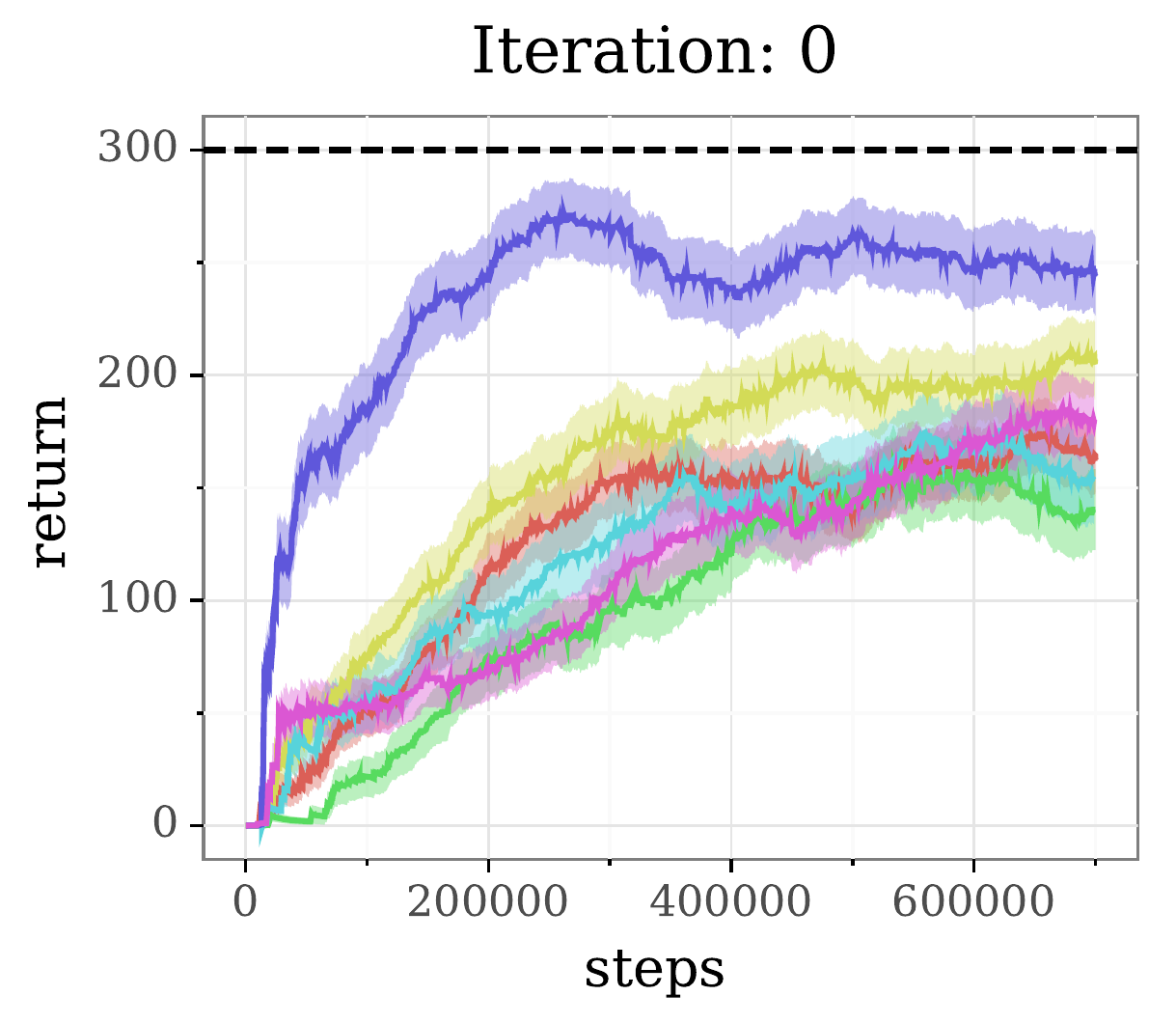}
  \includegraphics[height=0.145\textheight]{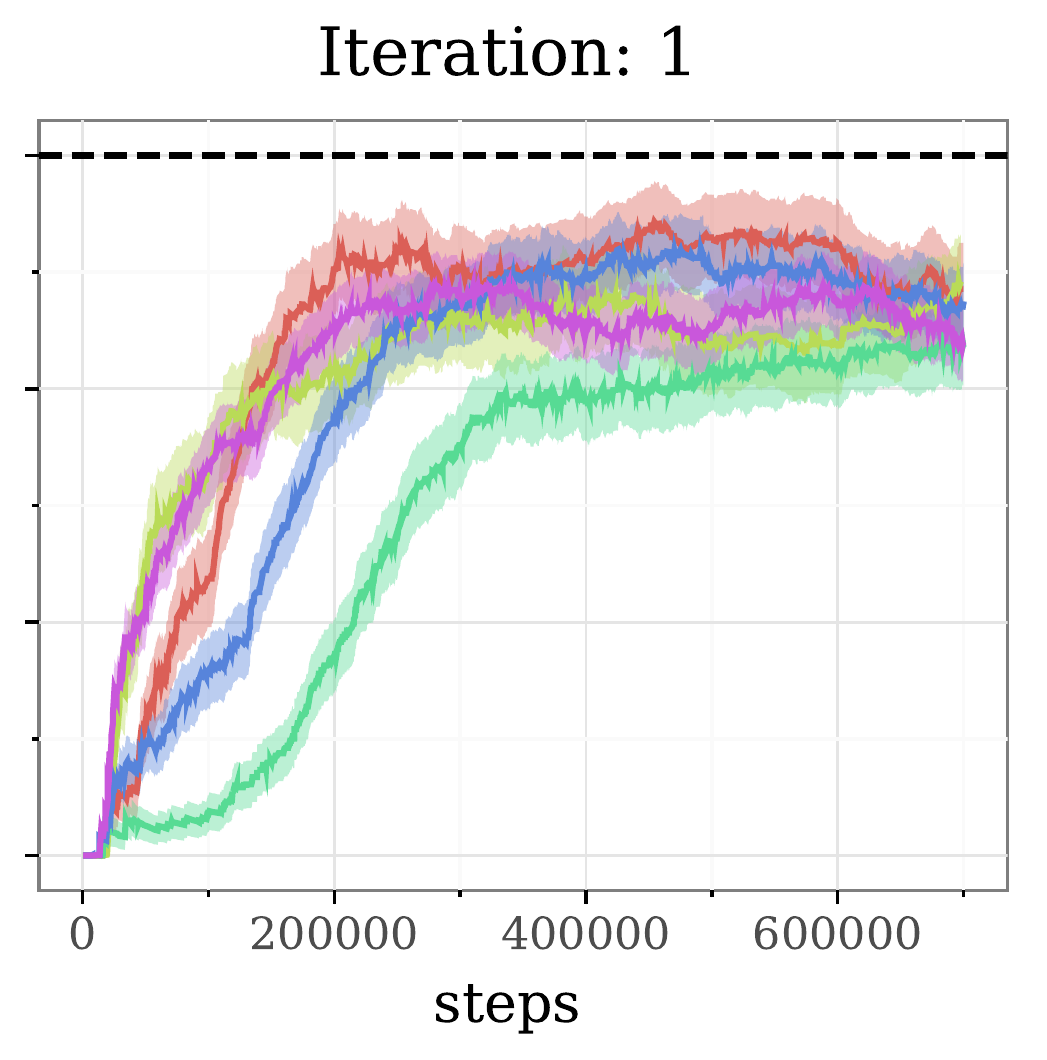}
  \includegraphics[height=0.145\textheight]{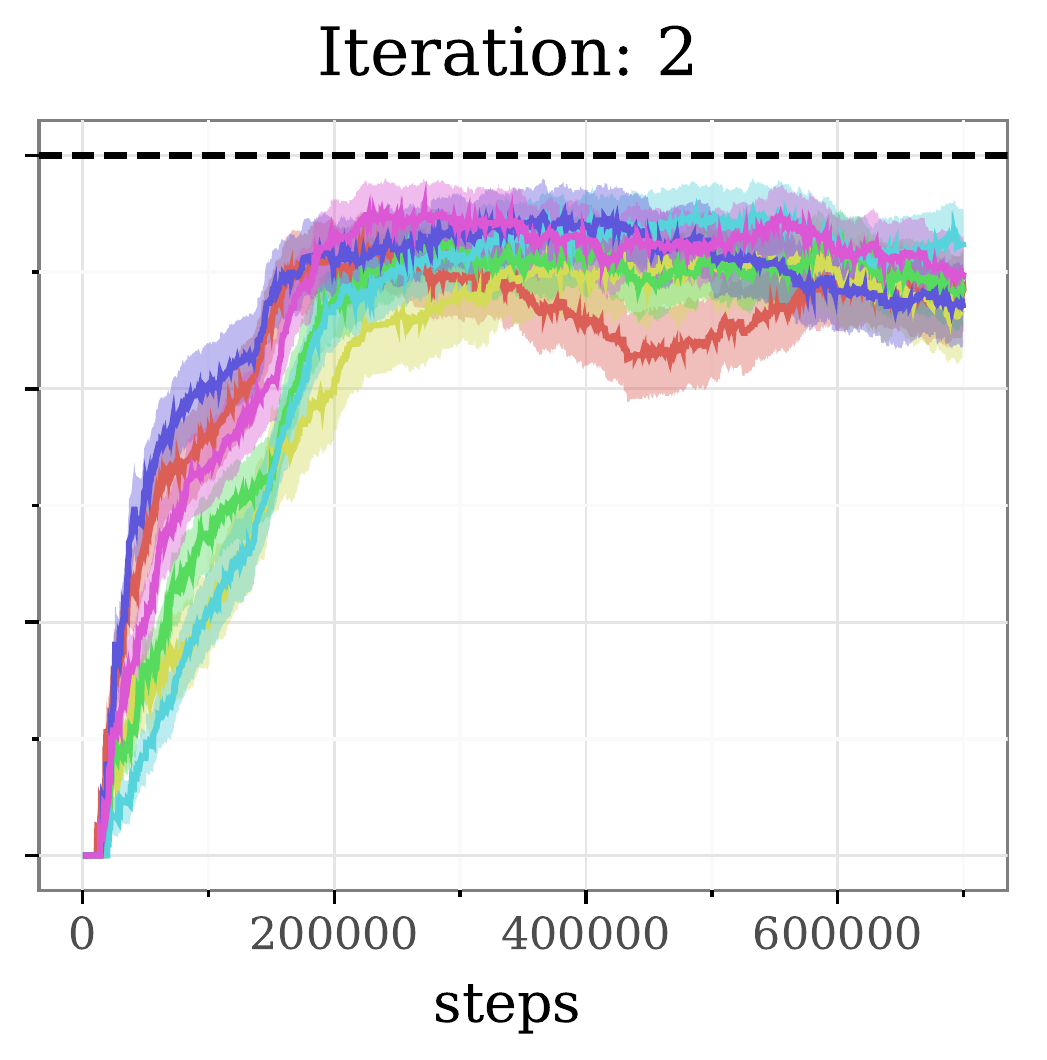}
  \includegraphics[height=0.145\textheight]{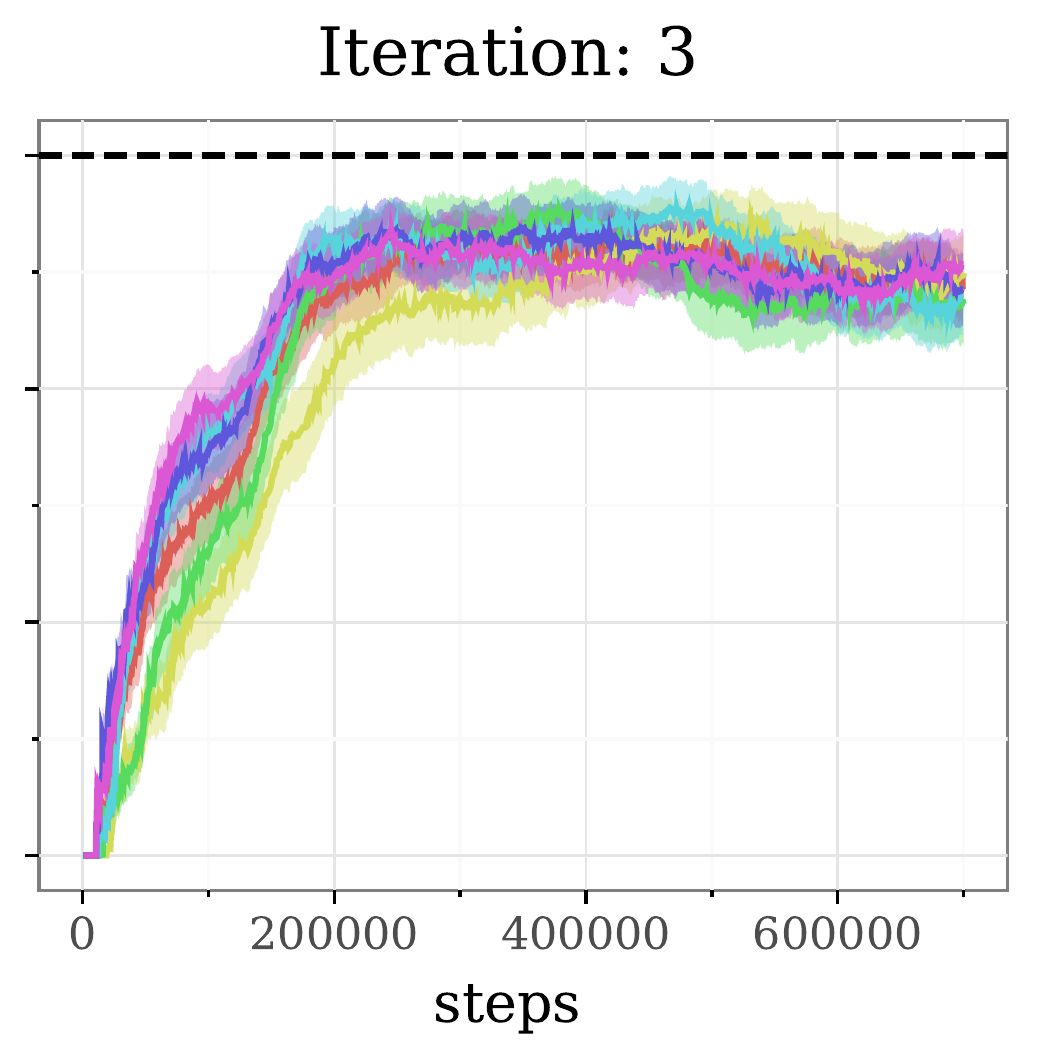}
  \caption{Policy training results for \textbf{TGR} reward models with different hyper-parameters (different colours of curves) as different iterations of refinement. The robustness of policy training increases with the iterations. Black dotted line indicated the performance of \textbf{GT}.}
  \label{fig:MIL-iterations}
\end{figure}

\autoref{fig:MIL-iterations} shows the improvement in \textbf{TGR-i} policies with the iterations of refinements.
Six training curves in the figure are selected according to different validation scores of the reward models and four figures correspond to iterations of reward model refinement.
We observe that while the performances of the best policies are not changing dramatically between iterations, the diversity of the returns decreases.
It means that even suboptimal policies perform well after refinement and thus the task of model selection is facilitated.

\section{Conclusions}

In this article we presented several approaches towards learning the reward functions for offline RL with limited reward supervision. 
As offline RL brings a promise of dealing with real-word tasks by relying on the pre-recorded datasets, it is an important step to consider the situations where the reward is not available as it may often occure in practice.
We studied two types of reward supervision, episode-level and timestep-level annotations.
We analysed the performance in terms of reward prediction quality and in terms of the quality of the policies trained with them. 
It turned out that using semi-supervised reward models can closely approach the policy performance with the ground truth rewards, both in settings with episode-level annotations and timestep-level annotations.  

\paragraph{Future directions}
We identified useful correlations between the reward model quality and policy performance, however, the dependency is not very strong and more studies might be needed. 
Reward model selection is closely related to offline policy evaluation in RL. 
A lot of progress in this area has been made recently~\cite{paine2020hyperparameter} and reward model learning can benefit from it.
So far, the proposed reward refinement procedure was applied to one strategy (\textbf{TGR}), but one can think of combining it with other reward learning algorithms (\emph{e.g.}, \textbf{ORIL}). 
We can also consider other types of annotations and their combinations.
Finally, the relation between the type of offline RL algorithm and the reward model quality is not yet completely understood (\autoref{fig:sup-d4pg}) and requires further investigation. 
\clearpage
{\small
\bibliographystyle{plainnat}
\bibliography{string,refs}

\begin{thebibliography}{39}
\providecommand{\natexlab}[1]{#1}
\providecommand{\url}[1]{\texttt{#1}}
\expandafter\ifx\csname urlstyle\endcsname\relax
  \providecommand{\doi}[1]{doi: #1}\else
  \providecommand{\doi}{doi: \begingroup \urlstyle{rm}\Url}\fi

\bibitem[Abbeel and Ng(2004)]{abbeel2004apprenticeship}
Pieter Abbeel and Andrew~Y Ng.
\newblock Apprenticeship learning via inverse reinforcement learning.
\newblock In \emph{International Conference on Machine Learning}, 2004.

\bibitem[Baram et~al.(2017)Baram, Anschel, Caspi, and Mannor]{baram2017end}
Nir Baram, Oron Anschel, Itai Caspi, and Shie Mannor.
\newblock End-to-end differentiable adversarial imitation learning.
\newblock In \emph{International Conference on Machine Learning}, 2017.

\bibitem[Bellemare et~al.(2017)Bellemare, Dabney, and
  Munos]{bellemare2017distributional}
Marc~G Bellemare, Will Dabney, and R{\'e}mi Munos.
\newblock A distributional perspective on reinforcement learning.
\newblock In \emph{International Conference on Machine Learning}, 2017.

\bibitem[Blum and Mitchell(1998)]{blum1998cotraining}
Avrim Blum and Tom Mitchell.
\newblock Combining labeled and unlabeled data with co-training.
\newblock In \emph{Conference on Computational Learning Theory}, 1998.

\bibitem[Cabi et~al.(2020)Cabi, G{\'o}mez~Colmenarejo, Novikov, Konyushkova,
  Reed, Jeong, Zolna, Aytar, Budden, Vecerik, Sushkov, Barker, Scholz, Denil,
  de~Freitas, and Wang]{cabi2020sketchy}
Serkan Cabi, Sergio G{\'o}mez~Colmenarejo, Alexander Novikov, Ksenia
  Konyushkova, Scott Reed, Rae Jeong, Konrad Zolna, Yusuf Aytar, David Budden,
  Mel Vecerik, Oleg Sushkov, David Barker, Jonathan Scholz, Misha Denil, Nando
  de~Freitas, and Ziyu Wang.
\newblock Scaling data-driven robotics with reward sketching and batch
  reinforcement learning.
\newblock In \emph{Robotics: Science and Systems Conference}, 2020.

\bibitem[Chen et~al.(2019)Chen, Zhou, Wang, Wang, Wu, Deng, and
  Ross]{chen2019bail}
Xinyue Chen, Zijian Zhou, Zheng Wang, Che Wang, Yanqiu Wu, Qing Deng, and Keith
  Ross.
\newblock {BAIL}: Best-action imitation learning for batch deep reinforcement
  learning.
\newblock arXiv:1910.12179, 2019.

\bibitem[Cinbis et~al.(2016)Cinbis, Verbeek, and Schmid]{cinbis2016mil}
Ramazan~Gokberk Cinbis, Jakob Verbeek, and Cordelia Schmid.
\newblock Weakly supervised object localization with multi-fold multiple
  instance learning.
\newblock In \emph{IEEE Transactions on Pattern Analysis and Machine
  Intelligence}, 2016.

\bibitem[Edwards et~al.(2016)Edwards, Isbell, and
  Takanishi]{edwards2016perceptual}
Ashley Edwards, Charles Isbell, and Atsuo Takanishi.
\newblock Perceptual reward functions.
\newblock In \emph{Deep Reinforcement Learning: Frontiers and Challenges
  Workshop at International Joint Conference on Artificial Intelligence}, 2016.

\bibitem[Elkan and Noto(2008)]{elkan2008learning}
Charles Elkan and Keith Noto.
\newblock Learning classifiers from only positive and unlabeled data.
\newblock In \emph{Conference on Knowledge Discovery and Data Mining}, 2008.

\bibitem[Finn et~al.(2016)Finn, Levine, and Abbeel]{finn2016guided}
Chelsea Finn, Sergey Levine, and Pieter Abbeel.
\newblock Guided cost learning: Deep inverse optimal control via policy
  optimization.
\newblock In \emph{International Conference on Machine Learning}, 2016.

\bibitem[Fu et~al.(2018)Fu, Luo, and Levine]{fu2017learning}
Justin Fu, Katie Luo, and Sergey Levine.
\newblock Learning robust rewards with adversarial inverse reinforcement
  learning.
\newblock In \emph{International Conference for Learning Representations},
  2018.

\bibitem[Fujimoto et~al.(2019)Fujimoto, Meger, and Precup]{fujimoto2019off}
Scott Fujimoto, David Meger, and Doina Precup.
\newblock Off-policy deep reinforcement learning without exploration.
\newblock In \emph{International Conference on Machine Learning}, 2019.

\bibitem[Ho and Ermon(2016)]{ho2016generative}
Jonathan Ho and Stefano Ermon.
\newblock Generative adversarial imitation learning.
\newblock In \emph{Advances in Neural Information Processing Systems}, 2016.

\bibitem[Hoffman et~al.(2020)Hoffman, Shahriari, Aslanides, Barth-Maron,
  Behbahani, Norman, Abdolmaleki, Cassirer, Yang, Baumli, Henderson, Novikov,
  Colmenarejo, Cabi, Gulcehre, Paine, Cowie, Wang, Piot, and
  de~Freitas]{hoffman2020acme}
Matt Hoffman, Bobak Shahriari, John Aslanides, Gabriel Barth-Maron, Feryal
  Behbahani, Tamara Norman, Abbas Abdolmaleki, Albin Cassirer, Fan Yang, Kate
  Baumli, Sarah Henderson, Alex Novikov, Sergio~Gómez Colmenarejo, Serkan
  Cabi, Caglar Gulcehre, Tom~Le Paine, Andrew Cowie, Ziyu Wang, Bilal Piot, and
  Nando de~Freitas.
\newblock Acme: A research framework for distributed reinforcement learning.
\newblock arXiv:2006.00979, 2020.

\bibitem[Kiryo et~al.(2017)Kiryo, Niu, Du~Plessis, and
  Sugiyama]{kiryo2017positive}
Ryuichi Kiryo, Gang Niu, Marthinus~C Du~Plessis, and Masashi Sugiyama.
\newblock Positive-unlabeled learning with non-negative risk estimator.
\newblock In \emph{Advances in Neural Information Processing Systems}, 2017.

\bibitem[Kotzias et~al.(2015)Kotzias, Denil, de~Freitas, and
  Smyth]{kotzias2015fromgroup}
Dimitrios Kotzias, Misha Denil, Nando de~Freitas, and Padhraic Smyth.
\newblock From group to individual labels using deep features.
\newblock In \emph{Conference on Knowledge Discovery and Data Mining}, 2015.

\bibitem[Lange et~al.(2012)Lange, Gabel, and Riedmiller]{lange2012batch}
Sascha Lange, Thomas Gabel, and Martin Riedmiller.
\newblock Batch reinforcement learning.
\newblock In \emph{Reinforcement Learning: State-of-the-Art}. Springer, 2012.

\bibitem[Levine et~al.(2020)Levine, Kumar, Tucker, and Fu]{levine2020offline}
Sergey Levine, Aviral Kumar, George Tucker, and Justin Fu.
\newblock Offline reinforcement learning: Tutorial, review, and perspectives on
  open problems.
\newblock arXiv:2005.01643, 2020.

\bibitem[Li et~al.(2017)Li, Song, and Ermon]{li2017infogail}
Yunzhu Li, Jiaming Song, and Stefano Ermon.
\newblock {InfoGAIL}: Interpretable imitation learning from visual
  demonstrations.
\newblock In \emph{Advances in Neural Information Processing Systems}, 2017.

\bibitem[Merel et~al.(2017)Merel, Tassa, Srinivasan, Lemmon, Wang, Wayne, and
  Heess]{merel2017learning}
Josh Merel, Yuval Tassa, Sriram Srinivasan, Jay Lemmon, Ziyu Wang, Greg Wayne,
  and Nicolas Heess.
\newblock Learning human behaviors from motion capture by adversarial
  imitation.
\newblock arXiv:1707.02201, 2017.

\bibitem[Ng and Russell(2000)]{ng2000algorithms}
Andrew~Y Ng and Stuart Russell.
\newblock Algorithms for inverse reinforcement learning.
\newblock In \emph{International Conference on Machine Learning}, 2000.

\bibitem[Osa et~al.(2018)Osa, Pajarinen, Neumann, Bagnell, Abbeel, and
  Peters]{osa2018an}
Takayuki Osa, Joni Pajarinen, Gerhard Neumann, J.~Andrew Bagnell, Pieter
  Abbeel, and Jan Peters.
\newblock \emph{An Algorithmic Perspective on Imitation Learning}.
\newblock Foundations and Trends in Robotics, 2018.

\bibitem[Paine et~al.(2020)Paine, Paduraru, Michi, Gulcehre, Zolna, Novikov,
  Wang, and de~Freitas]{paine2020hyperparameter}
Tom~Le Paine, Cosmin Paduraru, Andrea Michi, Caglar Gulcehre, Konrad Zolna,
  Alexander Novikov, Ziyu Wang, and Nando de~Freitas.
\newblock Hyperparameter selection for offline reinforcement learning.
\newblock ArXiv:2007.09055, 2020.

\bibitem[Peng et~al.(2019)Peng, Kumar, Zhang, and Levine]{peng2019advantage}
Xue~Bin Peng, Aviral Kumar, Grace Zhang, and Sergey Levine.
\newblock Advantage-weighted regression: Simple and scalable off-policy
  reinforcement learning.
\newblock arXiv:1910.00177, 2019.

\bibitem[Pomerleau(1989)]{pomerleau1989alvinn}
Dean~A Pomerleau.
\newblock {ALVINN}: An autonomous land vehicle in a neural network.
\newblock In \emph{Advances in Neural Information Processing Systems}, 1989.

\bibitem[Rahmatizadeh et~al.(2018)Rahmatizadeh, Abolghasemi, B{\"o}l{\"o}ni,
  and Levine]{rahmatizadeh2018vision}
Rouhollah Rahmatizadeh, Pooya Abolghasemi, Ladislau B{\"o}l{\"o}ni, and Sergey
  Levine.
\newblock Vision-based multi-task manipulation for inexpensive robots using
  end-to-end learning from demonstration.
\newblock In \emph{International Conference on Robotics and Automation}, 2018.

\bibitem[Reddy et~al.(2020)Reddy, Dragan, and Levine]{reddy2020squil}
Siddharth Reddy, Anca~D. Dragan, and Sergey Levine.
\newblock {SQIL}: Imitation learning via reinforcement learning with sparse
  rewards.
\newblock In \emph{International Conference for Learning Representations},
  2020.

\bibitem[Scudder(1965)]{scudder1965probability}
H.~Scudder.
\newblock Probability of error of some adaptive pattern-recognition machines.
\newblock In \emph{IEEE Transactions on Information Theory}, 1965.

\bibitem[Sermanet et~al.(2017)Sermanet, Xu, and
  Levine]{sermanet2017unsupervised}
Pierre Sermanet, Kelvin Xu, and Sergey Levine.
\newblock Unsupervised perceptual rewards for imitation learning.
\newblock In \emph{Robotics: Science and Systems Conference}, 2017.

\bibitem[Siegel et~al.(2020)Siegel, Springenberg, Berkenkamp, Abdolmaleki,
  Neunert, Lampe, Hafner, Heess, and Riedmiller]{siegel2020keep}
Noah Siegel, Jost~Tobias Springenberg, Felix Berkenkamp, Abbas Abdolmaleki,
  Michael Neunert, Thomas Lampe, Roland Hafner, Nicolas Heess, and Martin
  Riedmiller.
\newblock Keep doing what worked: Behavior modelling priors for offline
  reinforcement learning.
\newblock In \emph{International Conference for Learning Representations},
  2020.

\bibitem[Silver et~al.(2014)Silver, Lever, Heess, Degris, Wierstra, and
  Riedmiller]{silver2014deterministic}
David Silver, Guy Lever, Nicolas Heess, Thomas Degris, Daan Wierstra, and
  Martin Riedmiller.
\newblock Deterministic policy gradient algorithms.
\newblock In \emph{International Conference on Machine Learning}, 2014.

\bibitem[Singh et~al.(2019)Singh, Yang, Hartikainen, Finn, and
  Levine]{singh2019endtoend}
Avi Singh, Larry Yang, Kristian Hartikainen, Chelsea Finn, and Sergey Levine.
\newblock End-to-end robotic reinforcement learning without reward engineering.
\newblock In \emph{Robotics: Science and Systems Conference}, 2019.

\bibitem[Wang et~al.(2018)Wang, Xiong, Han, sun, Liu, and
  Zhang]{wang2018exponentially}
Qing Wang, Jiechao Xiong, Lei Han, peng sun, Han Liu, and Tong Zhang.
\newblock Exponentially weighted imitation learning for batched historical
  data.
\newblock In \emph{Advances in Neural Information Processing Systems}, 2018.

\bibitem[Wang et~al.(2020)Wang, Novikov, {\.Z}o{\l}na, Springenberg, Reed,
  Shahriari, Siegel, Merel, Gulcehre, Heess, and de~Freitas]{wang2020critic}
Ziyu Wang, Alexander Novikov, Konrad {\.Z}o{\l}na, Jost~Tobias Springenberg,
  Scott Reed, Bobak Shahriari, Noah Siegel, Josh Merel, Caglar Gulcehre,
  Nicolas Heess, and Nando de~Freitas.
\newblock Critic regularized regression.
\newblock In \emph{Advances in Neural Information Processing Systems}, 2020.

\bibitem[Xie et~al.(2020)Xie, Luong, Hovy, and Le]{xie2020selftraining}
Qizhe Xie, Minh-Thang Luong, Eduard Hovy, and Quoc~V Le.
\newblock Self-training with noisy student improves {ImageNet} classification.
\newblock In \emph{Conference on Computer Vision and Pattern Recognition},
  2020.

\bibitem[Xu and Denil(2019)]{xu2019positive}
Danfei Xu and Misha Denil.
\newblock Positive-unlabeled reward learning.
\newblock arXiv:1911.00459, 2019.

\bibitem[Zhu et~al.(2018)Zhu, Wang, Merel, Rusu, Erez, Cabi, Tunyasuvunakool,
  Kram{\'{a}}r, Hadsell, de~Freitas, and Heess]{zhu2018reinforcement}
Yuke Zhu, Ziyu Wang, Josh Merel, Andrei~A. Rusu, Tom Erez, Serkan Cabi, Saran
  Tunyasuvunakool, J{\'{a}}nos Kram{\'{a}}r, Raia Hadsell, Nando de~Freitas,
  and Nicolas Heess.
\newblock Reinforcement and imitation learning for diverse visuomotor skills.
\newblock In \emph{Robotics: Science and Systems Conference}, 2018.

\bibitem[Zolna et~al.(2019)Zolna, Reed, Novikov, Colmenarej, Budden, Cabi,
  Denil, de~Freitas, and Wang]{zolna2019task}
Konrad Zolna, Scott Reed, Alexander Novikov, Sergio~Gomez Colmenarej, David
  Budden, Serkan Cabi, Misha Denil, Nando de~Freitas, and Ziyu Wang.
\newblock Task-relevant adversarial imitation learning.
\newblock arXiv:1910.01077, 2019.

\bibitem[Zolna et~al.(2020)Zolna, Novikov, Konyushkova, Gulcehre, Wang, Aytar,
  Denil, de~Freitas, and Reeds]{ORIL}
Konrad Zolna, Alexander Novikov, Ksenia Konyushkova, Caglar Gulcehre, Ziyu
  Wang, Yusuf Aytar, Misha Denil, Nando de~Freitas, and Scott Reeds.
\newblock Offline learning from demonstrations and unlabeled experience.
\newblock arXiv:2011.13885, 2020.

\end{thebibliography}
}

\clearpage
\section{Supplementary materials}

\begin{figure}[h]
\centering
  \includegraphics[height=0.14\textheight]{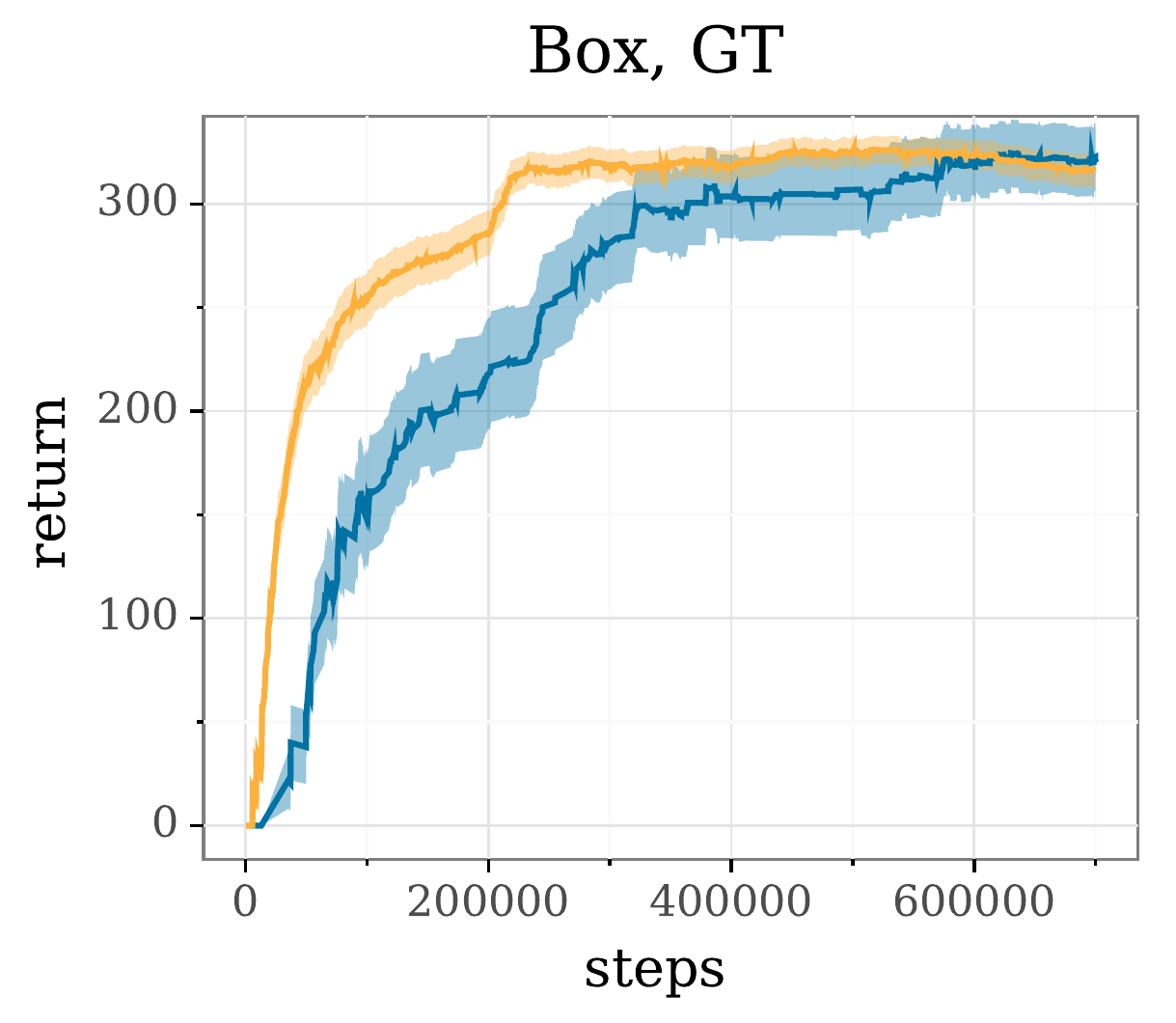}
  \includegraphics[height=0.14\textheight]{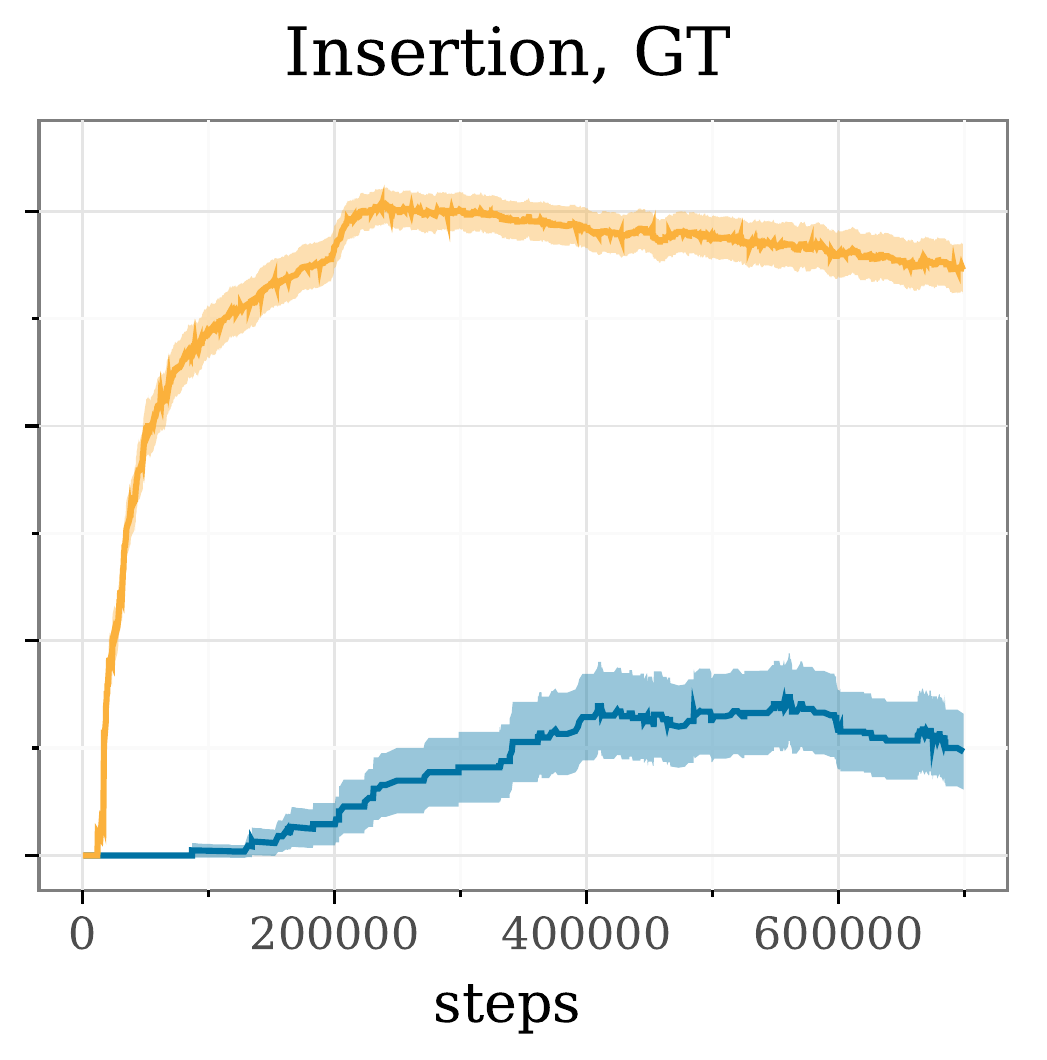}
  \includegraphics[height=0.14\textheight]{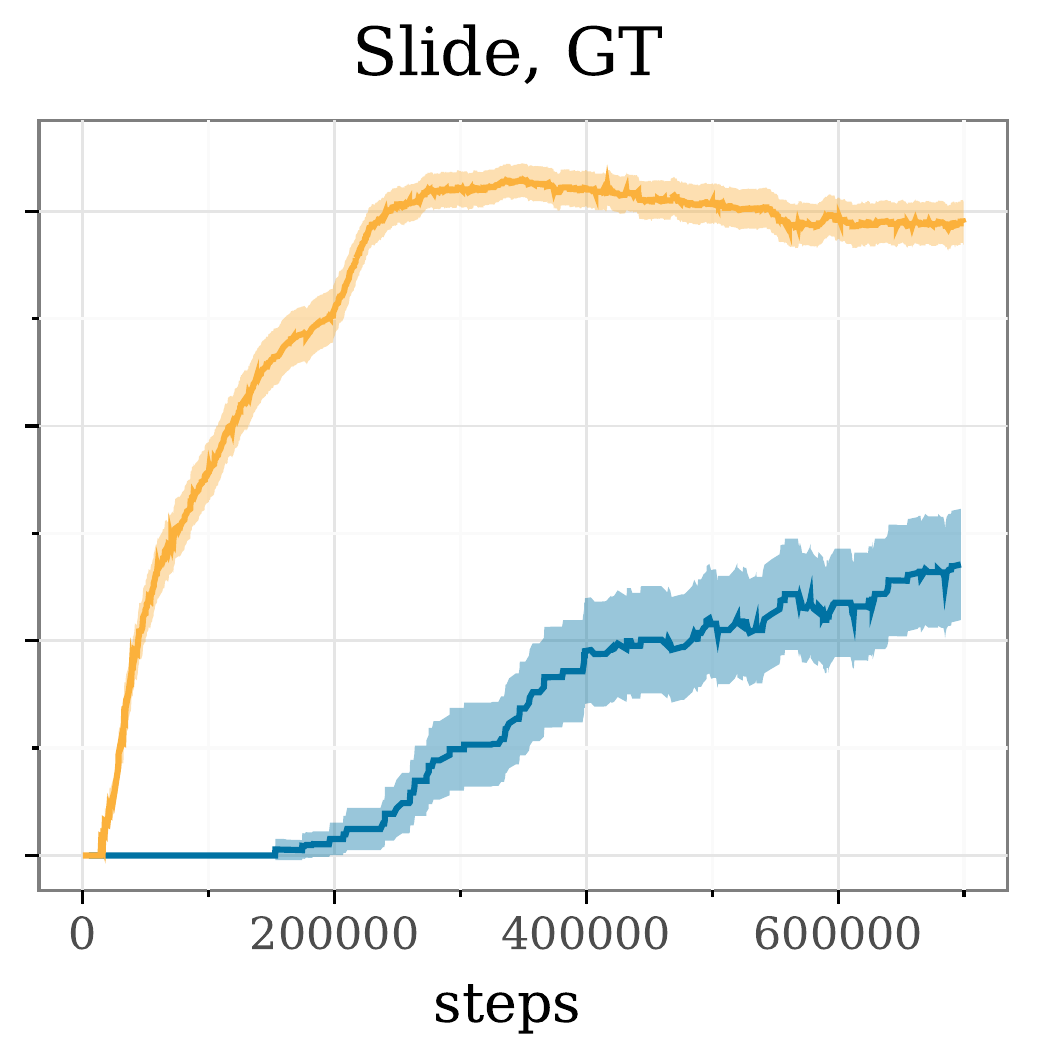}
  \includegraphics[height=0.14\textheight]{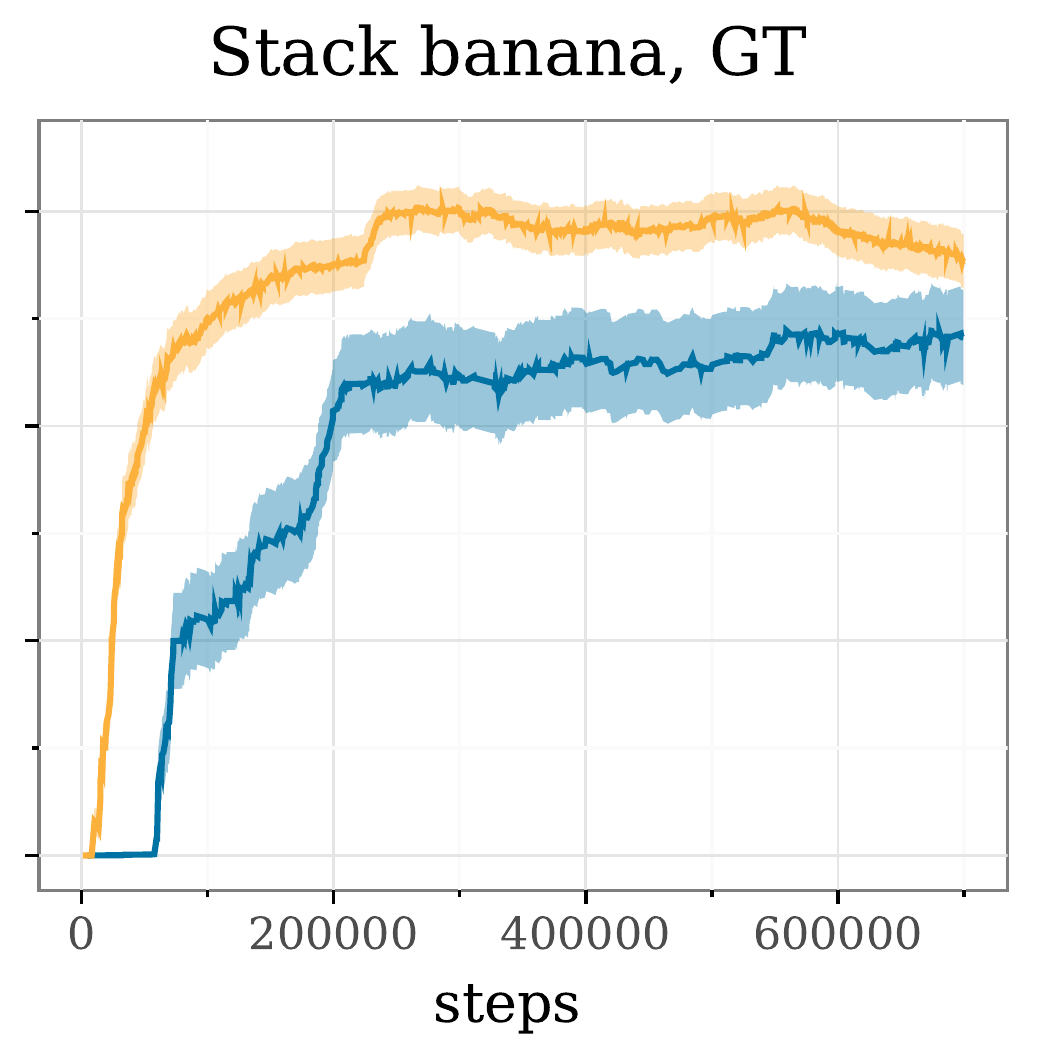}
  \includegraphics[height=0.14\textheight]{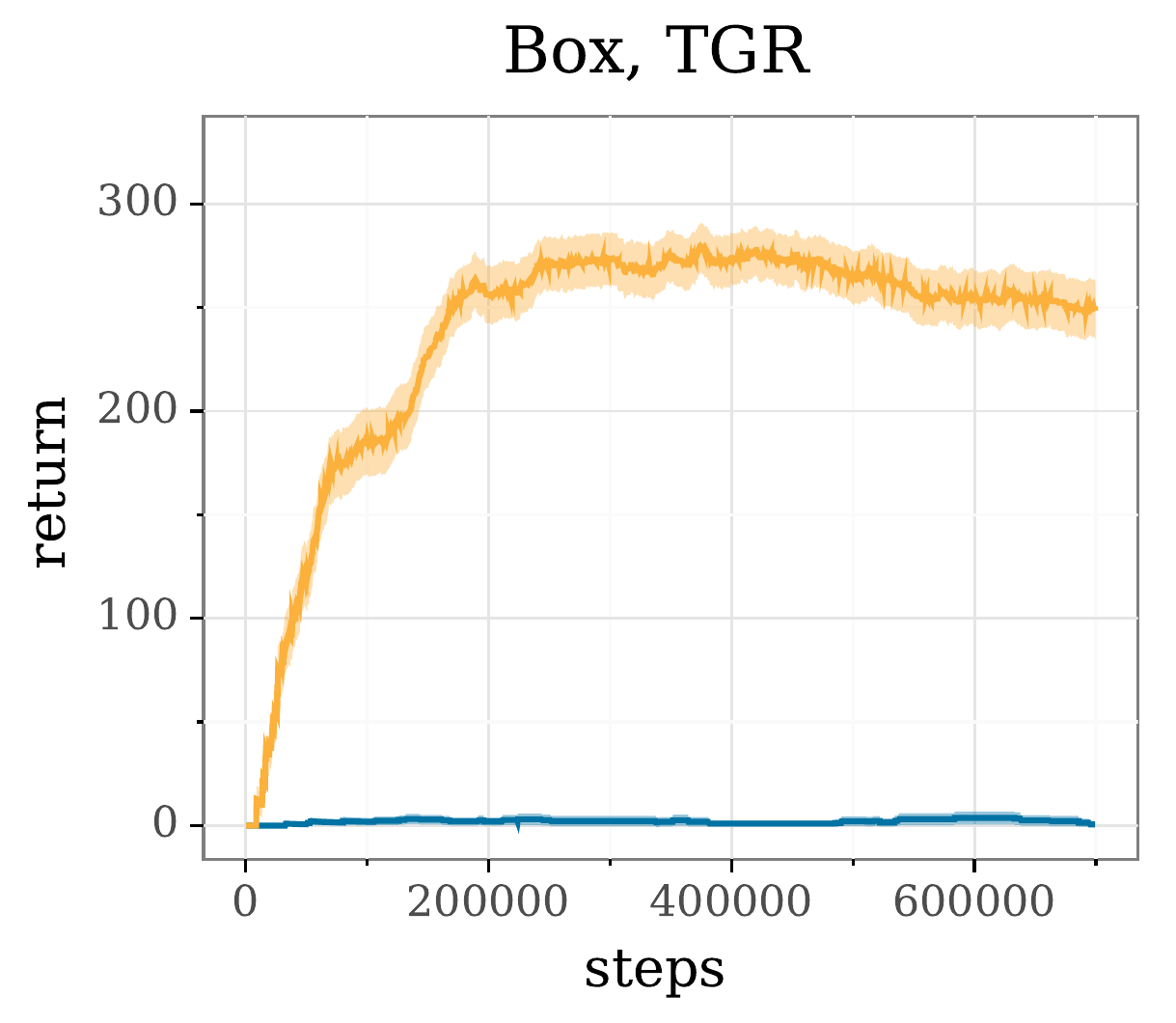}
  \includegraphics[height=0.14\textheight]{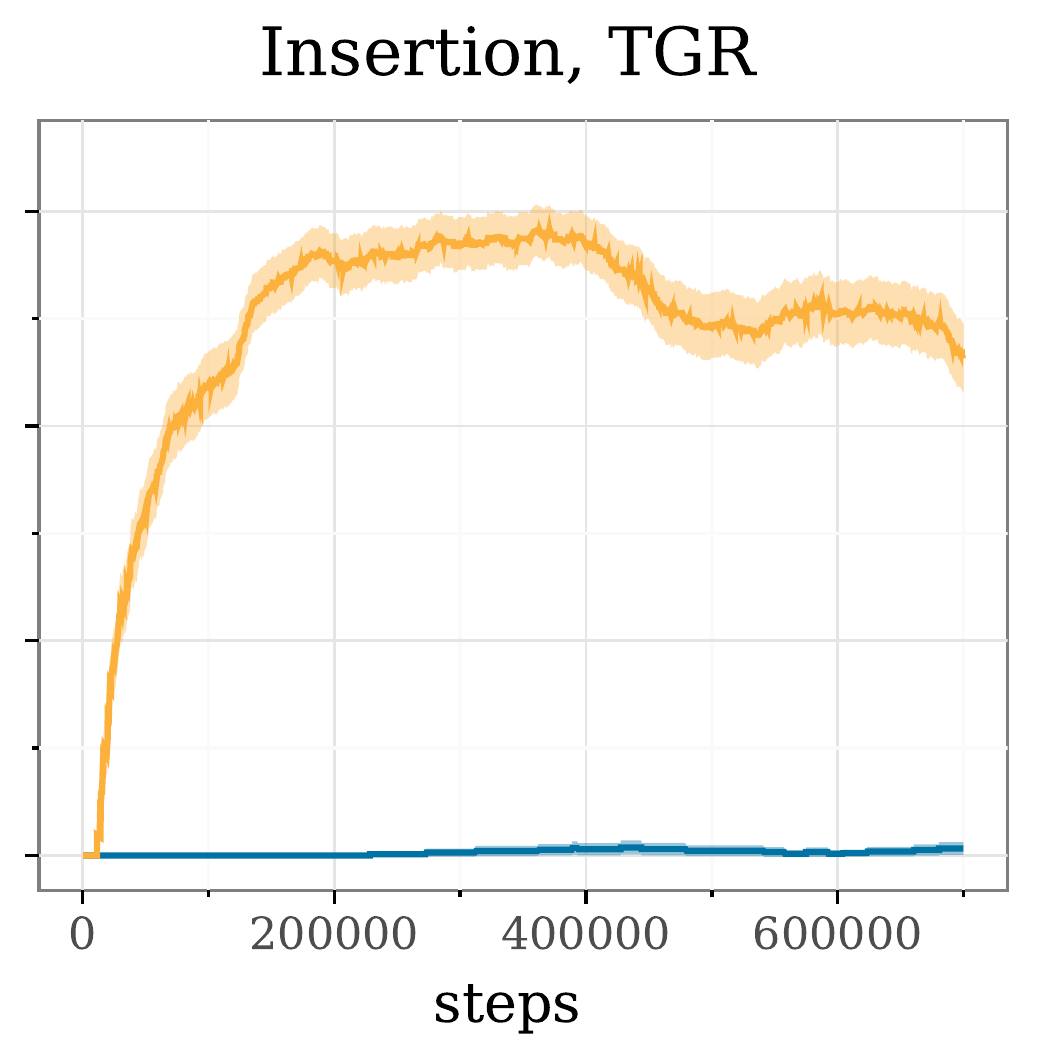}
  \includegraphics[height=0.14\textheight]{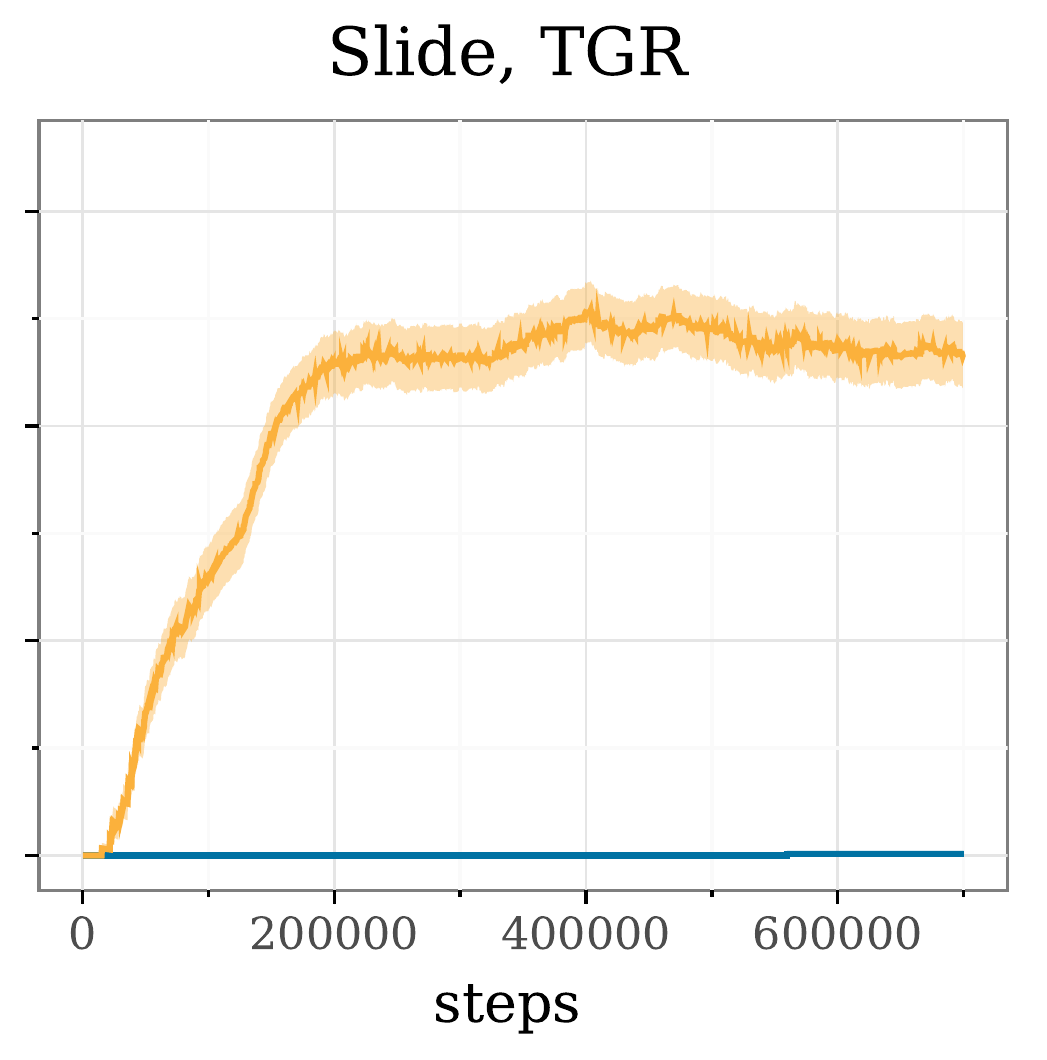}
  \includegraphics[height=0.14\textheight]{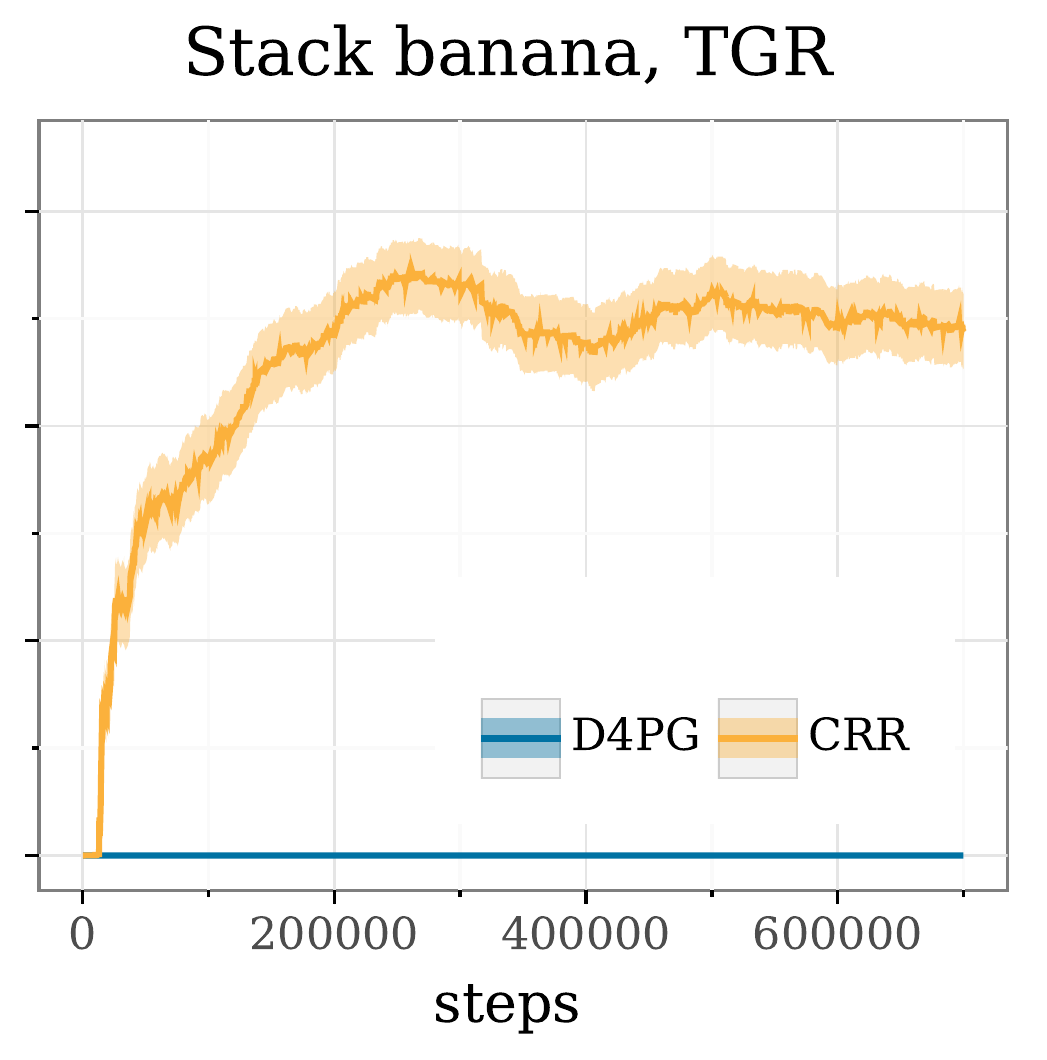}
  \caption{Results of policy training with two offline RL algorithms: CRR and D4PG for \num{4}. The first row shows the performance when using the ground truth reward signal and the second row is with a learnt reward function. The performance of CRR with a learnt reward function stays close to the \textbf{GT}, however, D4PG with learnt reward fails completely.}
  \label{fig:sup-d4pg}
\end{figure}

\begin{figure}[b]
\centering
  \includegraphics[height=0.125\textheight]{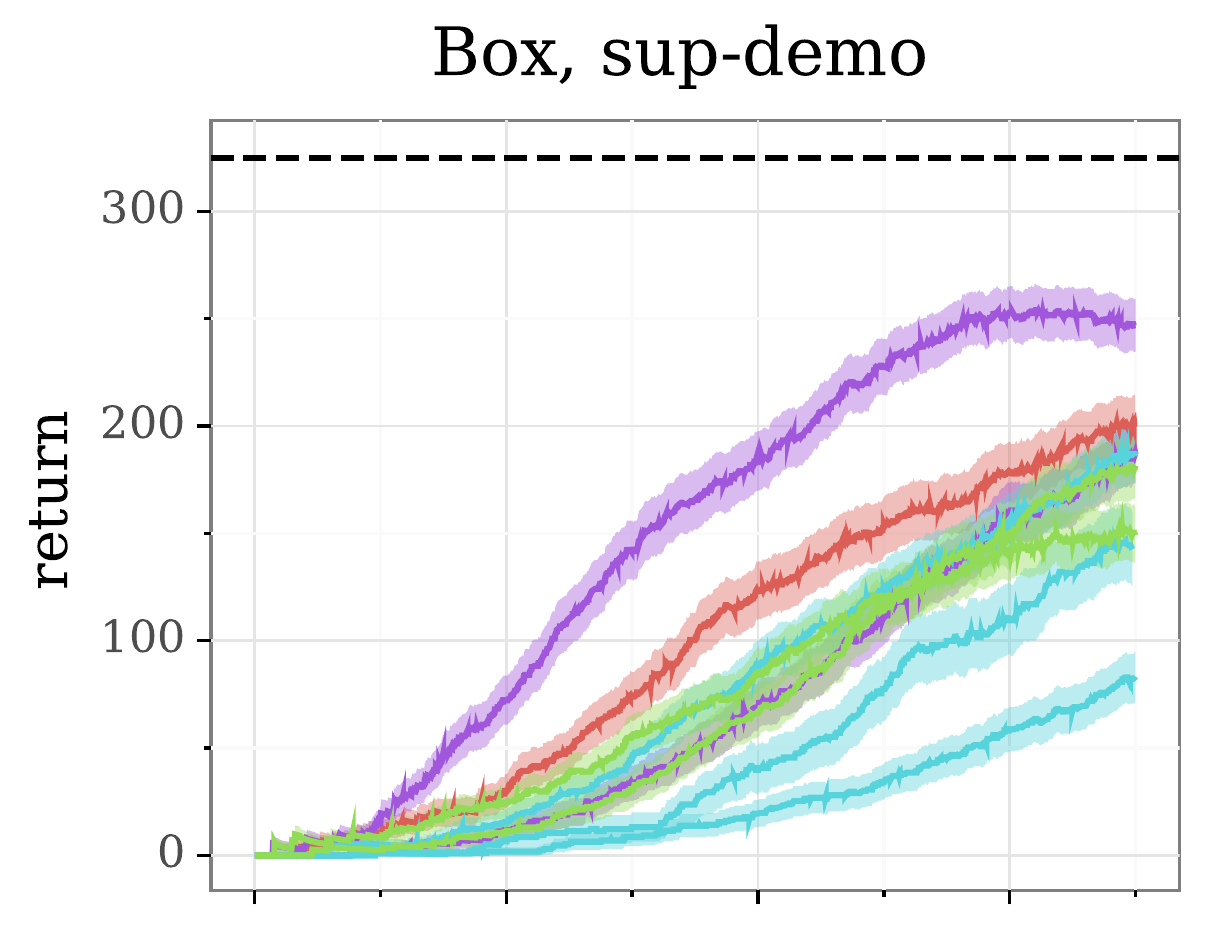}
  \includegraphics[height=0.125\textheight]{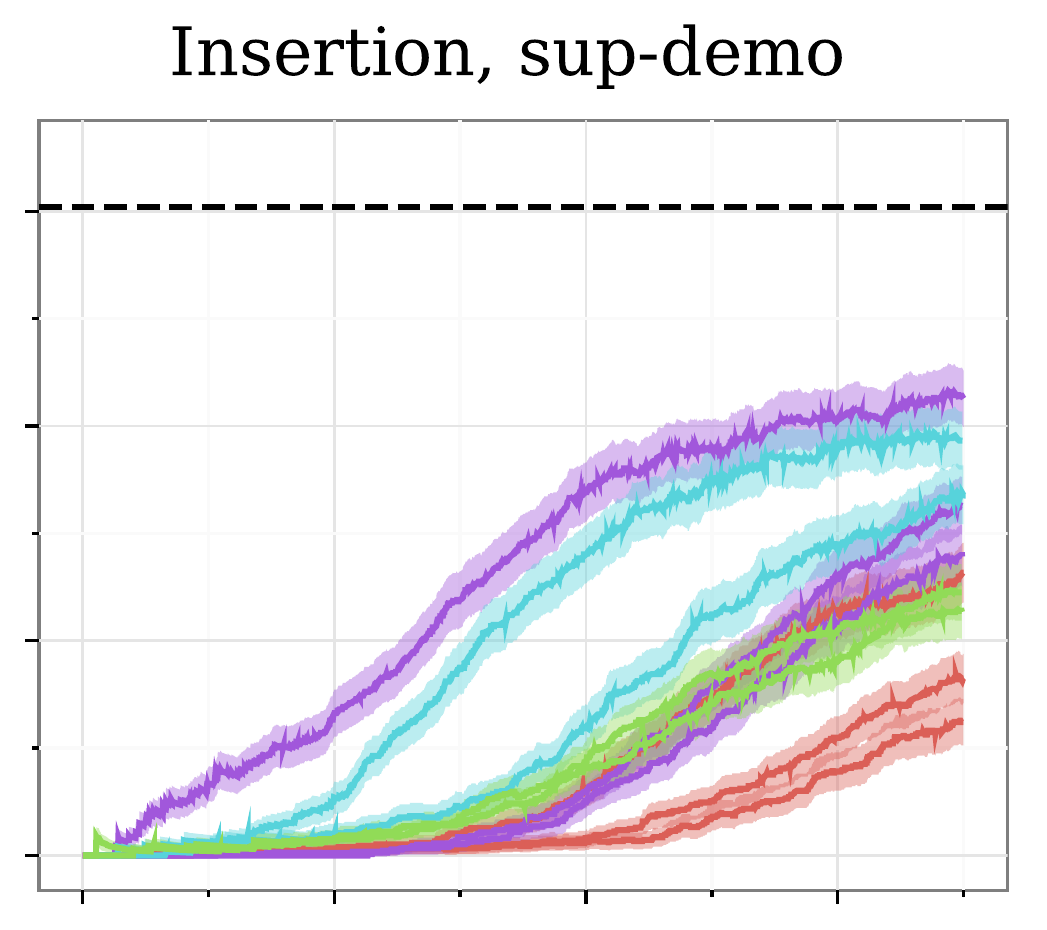}
  \includegraphics[height=0.125\textheight]{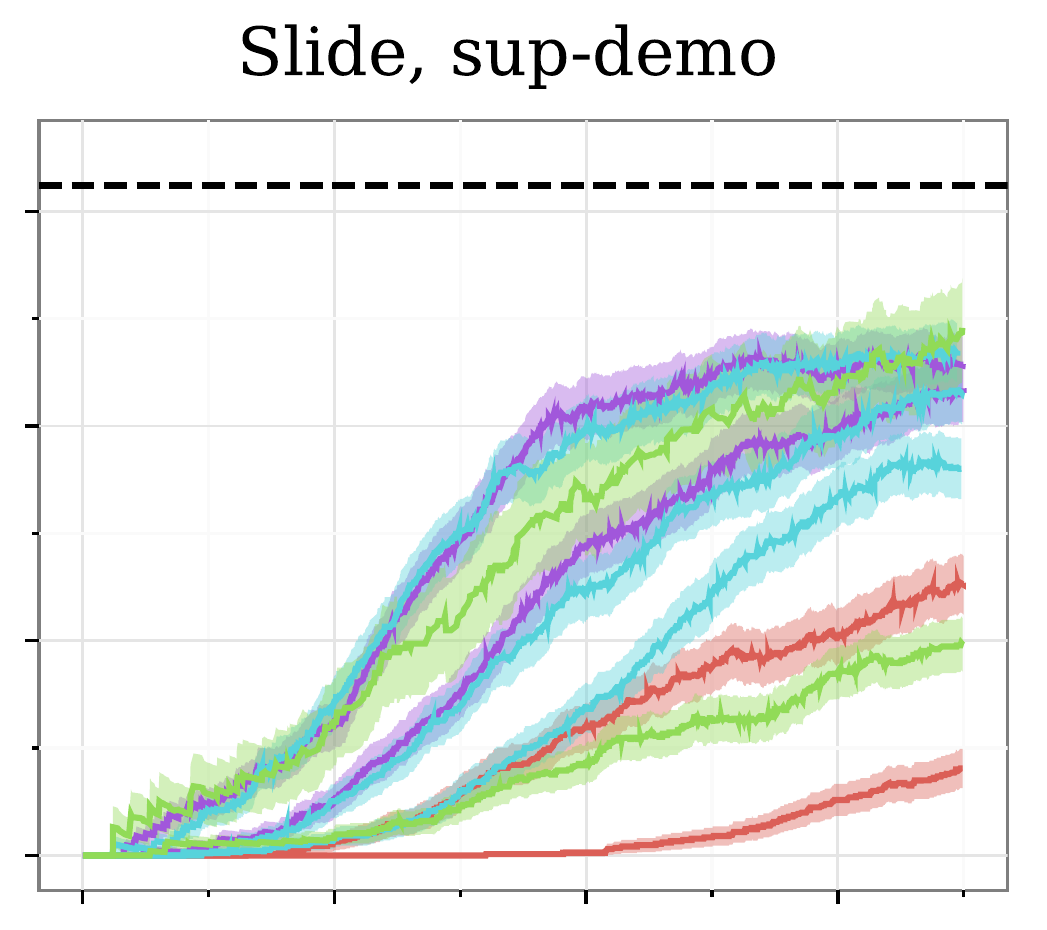}
  \includegraphics[height=0.125\textheight]{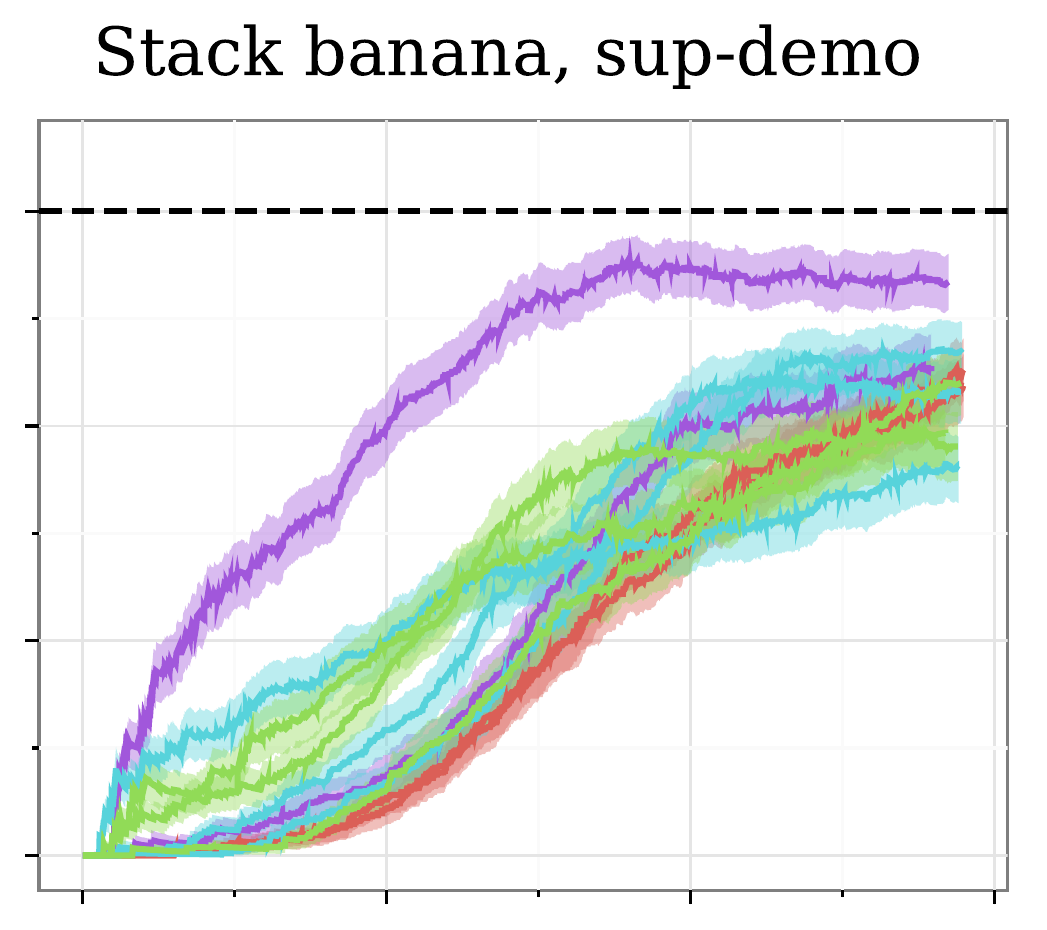}
  \includegraphics[height=0.14\textheight]{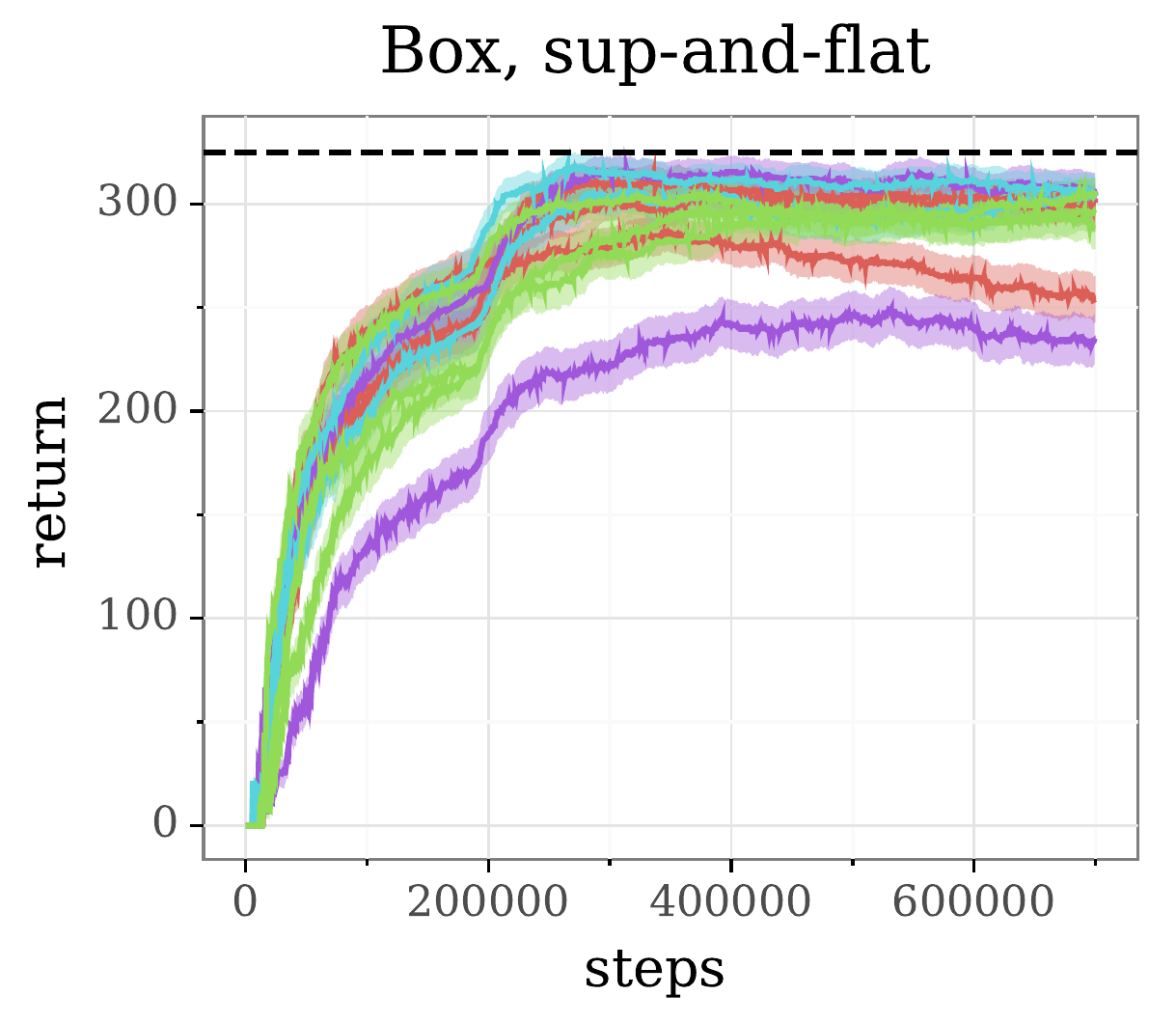}
  \includegraphics[height=0.14\textheight]{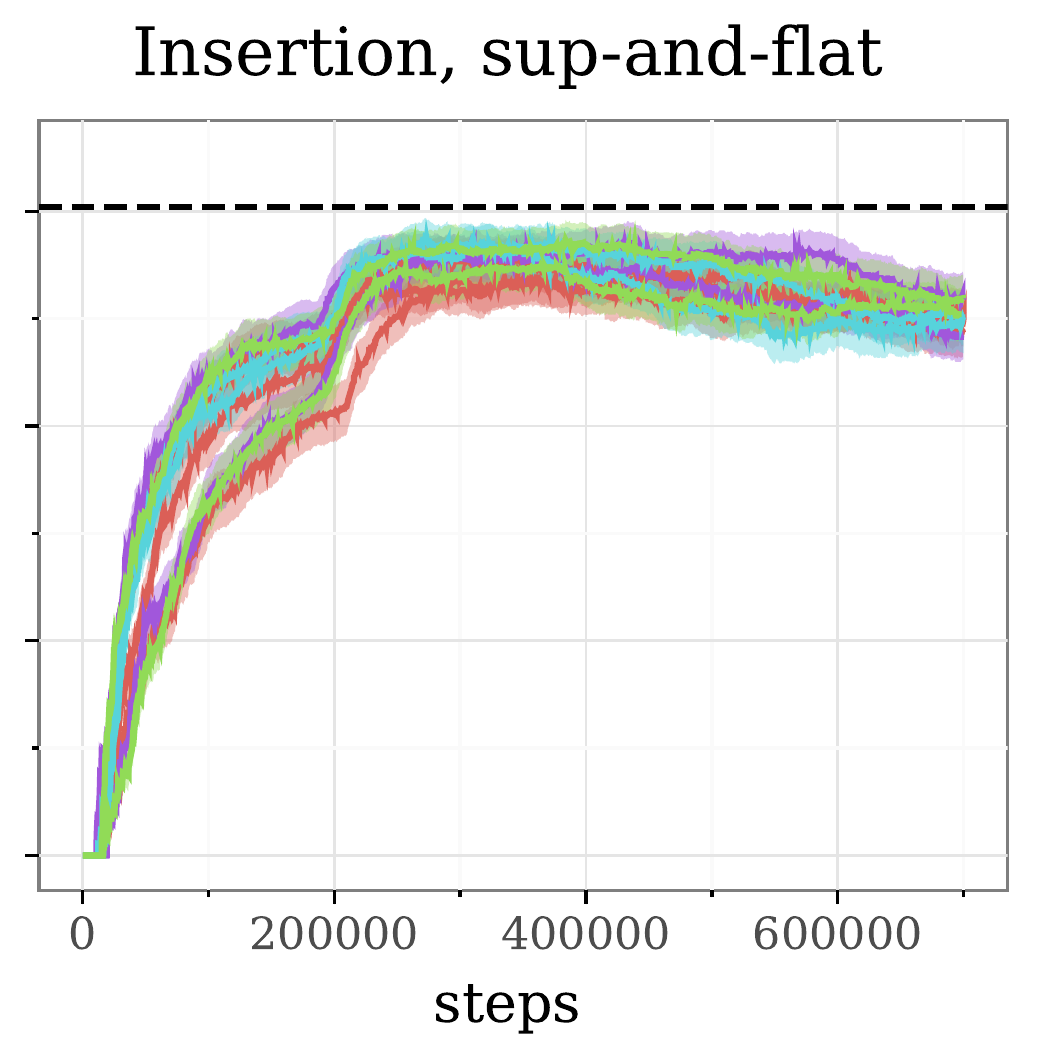}
  \includegraphics[height=0.14\textheight]{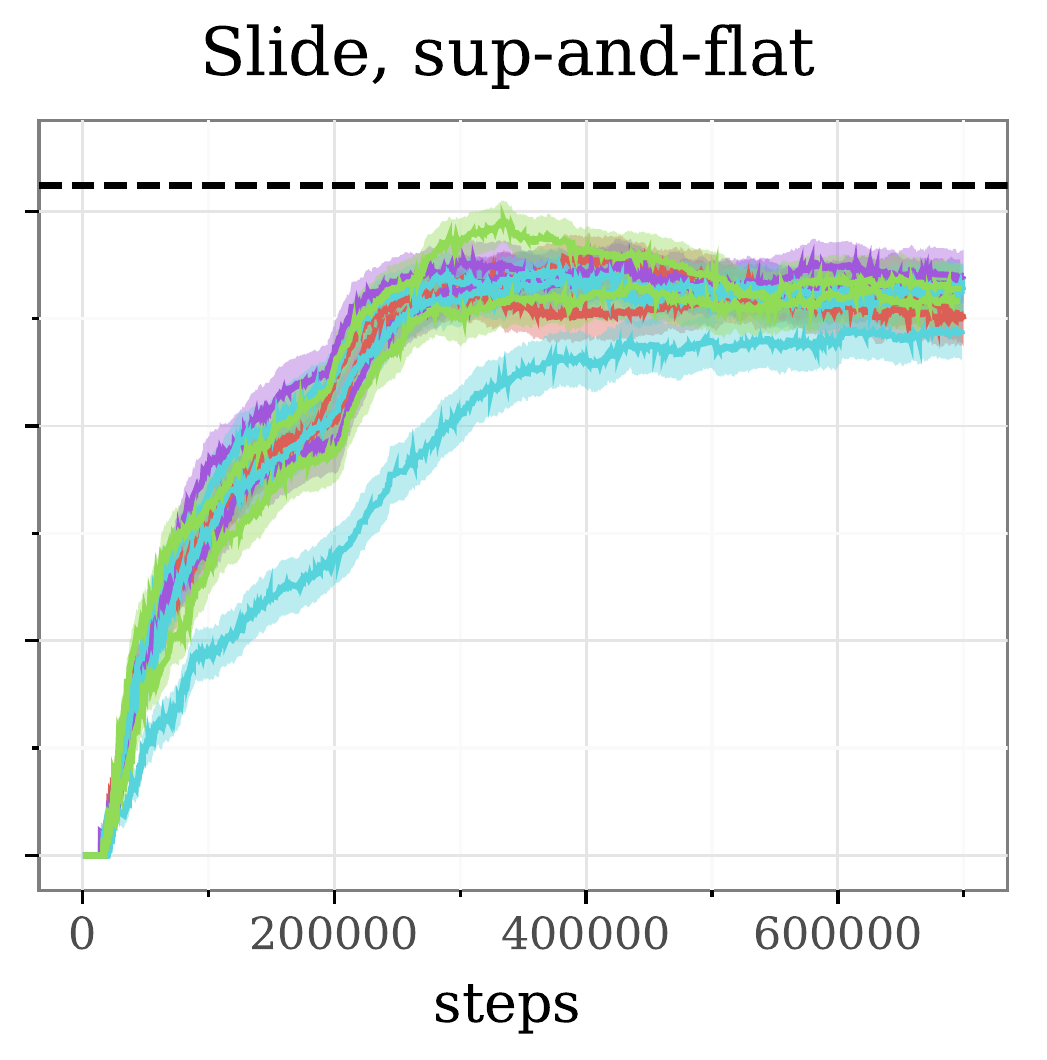}
  \includegraphics[height=0.14\textheight]{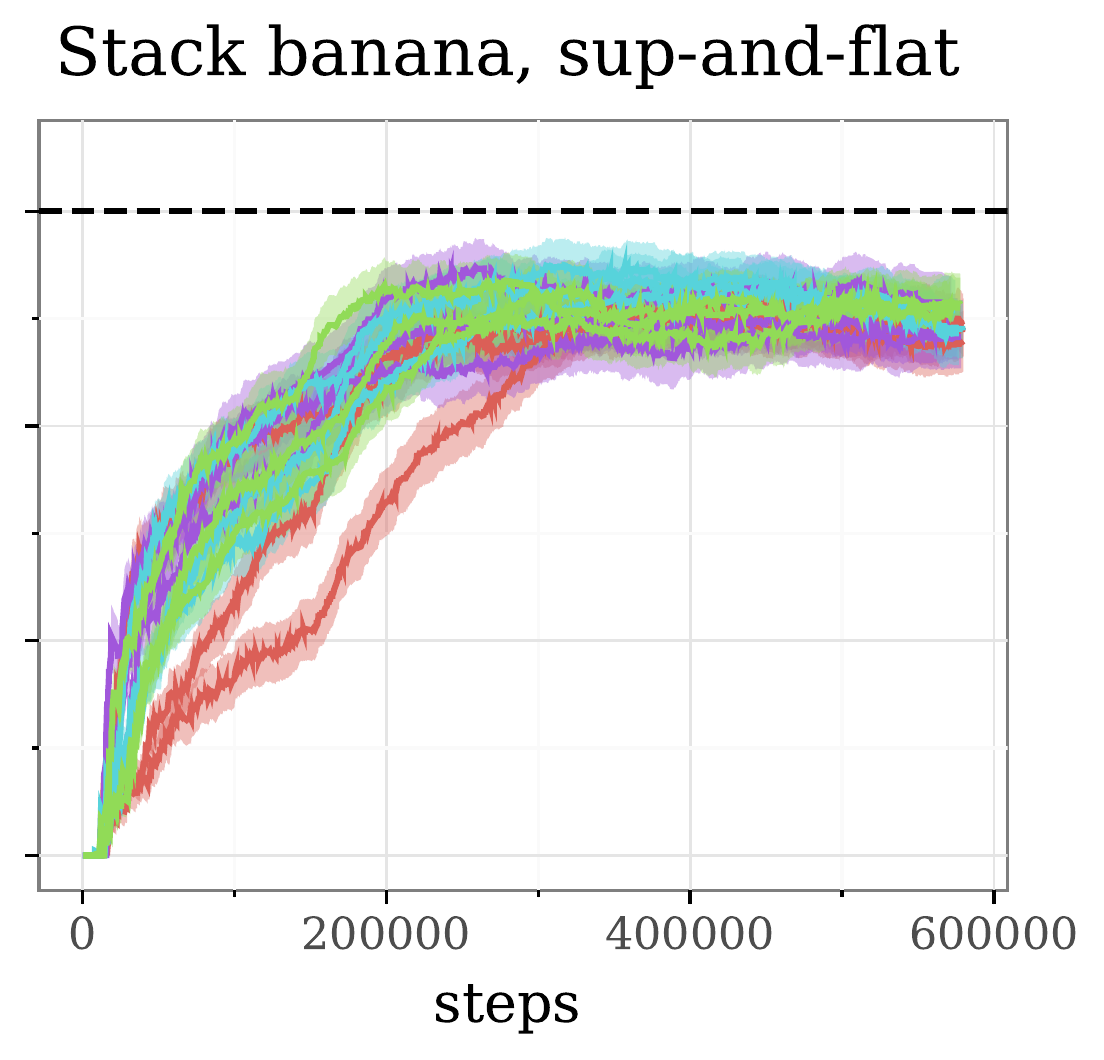}
  \includegraphics[width=0.3\linewidth]{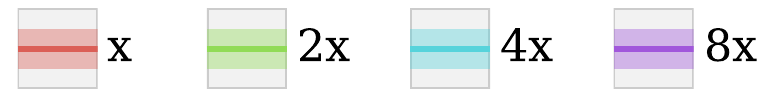}
  \caption{Results of policy training for \num{4} tasks using two reward model: \textbf{sup-demo} (first row) and \textbf{sup-and-flat} (second row) trained with varying amount of supervision. The performance of \textbf{sup-demo} can be potentially improved with more timesteps-level annotations, but the performance of \textbf{sup-and-flat} is almost indifferent to the amount of direct supervision and superior to \textbf{sup-demo}.}
  \label{fig:sup-moredata}
\end{figure}

\paragraph{Reward learning and different offline RL algorithms} As the first step towards understanding how the type of offline RL algorithm influences the performance of the model with learnt reward, we conduct the following experiment. 
We study the performance of D4PG algorithm~\cite{silver2014deterministic, bellemare2017distributional} (which was used in the related work by~\citet{cabi2020sketchy}) and CRR trained with ground truth rewards and learnt reward model.
The first row of \autoref{fig:sup-d4pg} shows the performance of CRR and D4PG when relying on the ground truth reward signal. 
Although CRR is clearly more efficient than D4PG, D4PG still reaches reasonable scores in two tasks and non-zero scores in two other tasks. 
The second row shows the performance of both algorithms with one of the learnt reward models.
While CRR still attains reasonably high scores in all tasks, D4PG struggles to learn any useful policy from this imprecise reward signal\footnote{Note that using D4PG with a simple classification reward model with episode-level annotations closely resembles the offline variant of GAIL.}. 
Our intuitive explanation of such difference in the behaviour of two offline algorithms is that CRR relies on the rewards only when learning a critic and it uses a loss similar to BC for acting.
This makes it more robust to the errors in reward predictions. 
However, further investigation is needed to understand how reward models inter-plays with the type of policy training.

\paragraph{Varying amount of timestep-level annotations}
Nest, we study at the performance of methods with timestep-level annotations with the increasing amount of supervision. 
\autoref{fig:sup-moredata} shows the performance of \textbf{sup-demo} in the first row and the second row shows the performance of \textbf{sup-and-flat}. 
We use growing amounts of data: 1) the same amount of data as in Sec.~\ref{sec:method_fromannotations} (\textbf{x}), 2) twice more (\textbf{2x}), 3) four times more (\textbf{4x}), and 4) eight times more (\textbf{8x}). 
The curves of the same colour use the same amount of data, but different hyperparameters as discussed in Sec.~\ref{exp-reward}. 
When we use only annotated data as in \textbf{sup-demo}, we notice that it is possible to improve the performance substantially by increasing the amount of supervision, but seldom up to the level of \textbf{sup-and-flat}. 
Contrary to this, the performance of the policy trained with \textbf{sup-and-flat} seems to be much less affected by the amount of timestep-level supervision and works equally well with \textbf{1x} and \textbf{8x} annotations. 
We conclude that \textbf{sup-and-flat} training algorithm is well suited for working with limited supervision. 

\end{document}